\documentclass[final]{cvpr}

\usepackage{times}
\usepackage{epsfig}
\usepackage{graphicx}
\usepackage{amsmath}
\usepackage{amssymb}

\usepackage[ruled, vlined]{algorithm2e}
\usepackage{diagbox}
\usepackage{slashbox}
\usepackage{multirow}
\usepackage[caption=false]{subfig}
\usepackage{xcolor}
\usepackage{subfiles}
\usepackage{amsmath}
\usepackage{graphicx}

\usepackage{amsthm}
\usepackage{amsmath,amsfonts,amssymb}
\theoremstyle{definition}

\DeclareMathOperator*{\argmin}{arg\,min}

\newcommand{\tb}[1]{\textbf{#1}}

\newcommand{\D}{\mathcal{D}}

\newcommand{\del}[1]{}
\newcommand{\norm}[1]{\left\lVert#1\right\rVert}
\makeatletter
\newcommand*{\rom}[1]{\expandafter\@slowromancap\romannumeral #1@}
\makeatother

\usepackage[pagebackref=false,breaklinks=true,colorlinks,bookmarks=false]{hyperref}

\def\confYear{CVPR 2021}
\begin{document}
%%%%%%%%% TITLE
\title{Model-Contrastive Federated Learning}

\author{Qinbin Li\\
National University of Singapore\\
{\tt\small qinbin@comp.nus.edu.sg}
\and
Bingsheng He\\
National University of Singapore\\
{\tt\small hebs@comp.nus.edu.sg}
\and
Dawn Song\\
UC Berkeley\\
{\tt \small dawnsong@berkeley.edu}
}

\maketitle
\pagestyle{empty}
\thispagestyle{empty}

%%%%%%%%% ABSTRACT
\begin{abstract}
Federated learning enables multiple parties to collaboratively train a machine learning model without communicating their local data. A key challenge in federated learning is to handle the heterogeneity of local data distribution across parties. Although many studies have been proposed to address this challenge, we find that they fail to achieve high performance in image datasets with deep learning models. In this paper, we propose MOON: model-contrastive federated learning. MOON is a simple and effective federated learning framework. The key idea of MOON is to utilize the similarity between model representations to correct the local training of individual parties, i.e., conducting contrastive learning in model-level. Our extensive experiments show that MOON significantly outperforms the other state-of-the-art federated learning algorithms on various image classification tasks.

\end{abstract}

\section{Introduction}
Deep learning is data hungry. Model training can benefit a lot from a large and representative dataset (\eg, ImageNet \cite{deng2009imagenet} and COCO~\cite{lin2014microsoft}). However, data are usually dispersed among different parties in practice (\eg, mobile devices and companies). Due to the increasing privacy concerns and data protection regulations \cite{voigt2017eu}, parties cannot send their private data to a centralized server to train a model. 

To address the above challenge, federated learning \cite{kairouz2019advances,yang2019federated,litian2019survey,li2019flsurvey} enables multiple parties to jointly learn a machine learning model without exchanging their local data. A popular federated learning algorithm is FedAvg~\cite{mcmahan2016communication}. In each round of FedAvg, the updated local models of the parties are transferred to the server, which further aggregates the local models to update the global model. The raw data is not exchanged during the learning process. Federated learning has emerged as an important machine learning area and attracted many research interests \cite{mcmahan2016communication,lifedprox,karimireddy2019scaffold,li2020practical,Wang2020Federated,dai2020federated,hu2020oarf,caldas2018leaf,chaoyanghe2020fedml}. Moreover, it has been applied in many applications such as medical imaging \cite{kaissis2020secure,kumar2020blockchain}, object detection \cite{liu2020fedvision}, and landmark classification \cite{hsu2020federated}.

A key challenge in federated learning is the heterogeneity of data distribution on different parties \cite{kairouz2019advances}. The data can be non-identically distributed among the parties in many real-world applications, which can degrade the performance of federated learning~\cite{karimireddy2019scaffold,Li2020On,niidbench}. When each party updates its local model, its local objective may be far from the global objective. Thus, the averaged global model is away from the global optima. There have been some studies trying to address the non-IID issue in the local training phase~\cite{lifedprox,karimireddy2019scaffold}. FedProx \cite{lifedprox} directly limits the local updates by $\ell_2$-norm distance, while SCAFFOLD \cite{karimireddy2019scaffold} corrects the local updates via variance reduction \cite{johnson2013accelerating}. However, as we show in the experiments (see Section \ref{sec:exp}), these approaches fail to achieve good performance on image datasets with deep learning models, which can be as bad as FedAvg.

In this work, we address the non-IID issue from a novel perspective based on an intuitive observation: \emph{the global model trained on a whole dataset is able to learn a better representation than the local model trained on a skewed subset}. Specifically, we propose \tb{mo}del-c\tb{on}trastive learning (MOON), which corrects the local updates by maximizing the agreement of representation learned by the current local model and the representation learned by the global model. Unlike the traditional contrastive learning approaches \cite{simclr,simclr2,he2020momentum,misra2020self}, which achieve state-of-the-art results on learning visual representations by comparing the representations of different images, MOON conducts contrastive learning in model-level by comparing the representations learned by different models. Overall, MOON is a simple and effective federated learning framework, and addresses the non-IID data issue with the novel design of model-based contrastive learning.

We conduct extensive experiments to evaluate the effectiveness of MOON. MOON significantly outperforms the other state-of-the-art federated learning algorithms~\cite{mcmahan2016communication,lifedprox,karimireddy2019scaffold} on various image classification datasets including CIFAR-10, CIFAR-100, and Tiny-Imagenet. With only lightweight modifications to FedAvg, MOON outperforms existing approaches by at least 2\% accuracy in most cases. Moreover, the improvement of MOON is very significant on some settings. For example, on CIFAR-100 dataset with 100 parties, MOON achieves 61.8\% top-1 accuracy, while the best top-1 accuracy of existing studies is 55\%. 

\section{Background and Related Work}

\subsection{Federated Learning}
\begin{figure}
    \centering
    \includegraphics[width=\linewidth]{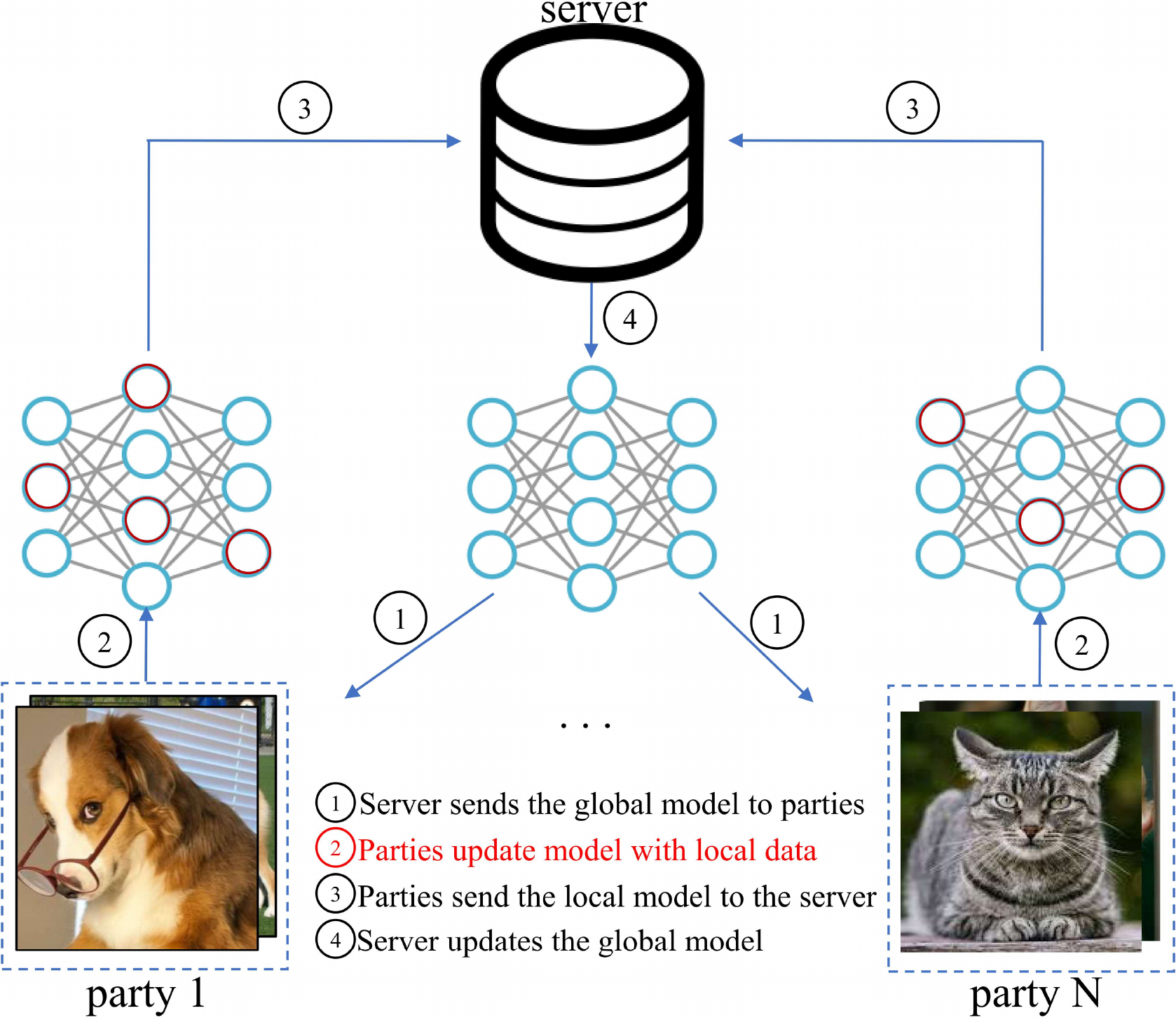}
    \caption{The FedAvg framework. In this paper, we focus on the second step, i.e., the local training phase.}
    \label{fig:fedavg_fram}
    \vspace{-5pt}
\end{figure}

FedAvg \cite{mcmahan2016communication} has been a de facto approach for federated learning. The framework of FedAvg is shown in Figure \ref{fig:fedavg_fram}. There are four steps in each round of FedAvg. First, the server sends a global model to the parties. Second, the parties perform stochastic gradient descent (SGD) to update their models locally. Third, the local models are sent to a central server. Last, the server averages the model weights to produce a global model for the training of the next round.

There have been quite some studies trying to improve FedAvg on non-IID data. Those studies can be divided into two categories: improvement on local training (i.e., step 2 of Figure \ref{fig:fedavg_fram}) and on aggregation (i.e., step 4 of Figure \ref{fig:fedavg_fram}). This study belongs to the first category.

As for studies on improving local training, FedProx \cite{lifedprox} introduces a proximal term into the objective during local training. The proximal term is computed based on the $\ell_2$-norm distance between the current global model and the local model. Thus, the local model update is limited by the proximal term during the local training. SCAFFOLD \cite{karimireddy2019scaffold} corrects the local updates by introducing control variates. Like the training model, the control variates are also updated by each party during local training. The difference between the local control variate and the global control variate is used to correct the gradients in local training. However, FedProx shows experiments on MNIST and EMNIST only with multinomial logistic regression, while SCAFFOLD only shows experiments on EMNIST with logistic regression and 2-layer fully connected layer. The effectiveness of FedProx and SCAFFOLD on image datasets with deep learning models has not been well explored. As shown in our experiments, those studies have little or even no advantage over FedAvg, which motivates this study for a new approach of handling non-IID image datasets with deep learning models. We also notice that there are other related contemporary work \cite{acar2021federated,li2021fedbn,wang2020addressing} when preparing this paper. We leave the comparison between MOON and these contemporary work as future studies.

As for studies on improving the aggregation phase, FedMA \cite{Wang2020Federated} utilizes Bayesian non-parametric methods to match and average weights in a layer-wise manner. FedAvgM \cite{hsu2019measuring} applies momentum when updating the global model on the server. Another recent study, FedNova \cite{wang2020tackling}, normalizes the local updates before averaging.
Our study is orthogonal to them and potentially can be combined with these techniques as we work on the local training phase.

Another research direction is personalized federated learning \cite{fallah2020personalized,dinh2020personalized,hanzely2020lower,zhang2021personalized,huang2021personalized}, which tries to learn personalized local models for each party. In this paper, we study the typical federated learning, which tries to learn a single global model for all parties.

\subsection{Contrastive Learning}

Self-supervised learning \cite{jing2020self,grill2020bootstrap,simclr,simclr2,he2020momentum,misra2020self} is a recent hot research direction, which tries to learn good data representations from unlabeled data. Among those studies, contrastive learning approaches \cite{simclr,simclr2,he2020momentum,misra2020self} achieve state-of-the-art results on learning visual representations. The key idea of contrastive learning is to reduce the distance between the representations of different augmented views of the same image (i.e., \emph{positive pairs}), and increase the distance between the representations of augmented views of different images (i.e., \emph{negative pairs}).

A typical contrastive learning framework is SimCLR \cite{simclr}. Given an image $x$, SimCLR first creates two correlated views of this image using different data augmentation operators, denoted $x_i$ and $x_j$. A base encoder $f(\cdot)$ and a projection head $g(\cdot)$ are trained to extract the representation vectors and map the representations to a latent space, respectively. Then, a contrastive loss (i.e., NT-Xent \cite{sohn2016improved}) is applied on the projected vector $g(f(\cdot))$, which tries to maximize agreement between differently augmented views of the same image. Specifically, given $2N$ augmented views and a pair of view $x_i$ and $x_j$ of same image, the contrastive loss for this pair is defined as 

\begin{equation}
l_{i, j} = -\log{\frac{\exp(\textrm{sim}(x_i, x_j)/\tau)}{\sum_{k=1}^{2N} \mathbb{I}_{[k\neq i]}\exp(\textrm{sim}(x_i, x_k)/\tau)}}
\end{equation}

where $\textrm{sim}(\cdot,\cdot)$ is a cosine similarity function and $\tau$ denotes a temperature parameter. The final loss is computed by summing the contrastive loss of all pairs of the same image in a mini-batch. 

Besides SimCLR, there are also other contrastive learning frameworks such as CPC \cite{oord2018representation}, CMC \cite{tian2019contrastive} and MoCo \cite{he2020momentum}. We choose SimCLR for its simplicity and effectiveness in many computer vision tasks. Still, the basic idea of contrastive learning is similar among these studies: the representations obtained from different images should be far from each other and the representations obtained from the same image should be related to each other. The idea is intuitive and has been shown to be effective. 

There is one recent study \cite{zhang2020federated} that combines federated learning with contrastive learning. They focus on the unsupervised learning setting. Like SimCLR, they use contrastive loss to compare the representations of different images. In this paper, we focus on the supervised learning setting and propose model-contrastive learning to compare representations learned by different models.
\section{Model-Contrastive Federated Learning}

\subsection{Problem Statement}
Suppose there are $N$ parties, denoted $P_1, ..., P_N$. Party $P_i$ has a local dataset $\D^i$. Our goal is to learn a machine learning model $w$ over the dataset $\D\triangleq \bigcup_{i\in[N]}\D^i$ with the help of a central server, while the raw data are not exchanged. The objective is to solve 
\begin{equation}
    \argmin_{w} \mathcal{L}(w) = \sum_{i=1}^N \frac{|\D^i|}{|\D|}L_i(w),
\end{equation}
where $L_i(w) = \mathbb{E}_{(x,y)\sim \D^i} [\ell_i(w; (x, y))]$ is the empirical loss of $P_i$.

\subsection{Motivation}

MOON is based on an intuitive idea: the model trained on the whole dataset is able to extract a better feature representation than the model trained on a skewed subset. For example, given a model trained on dog and cat images, we cannot expect the features learned by the model to distinguish birds and frogs which never exist during training. 

\begin{figure}
\centering
\subfloat[global model\label{fig:global}]{\includegraphics[width=0.5\linewidth]{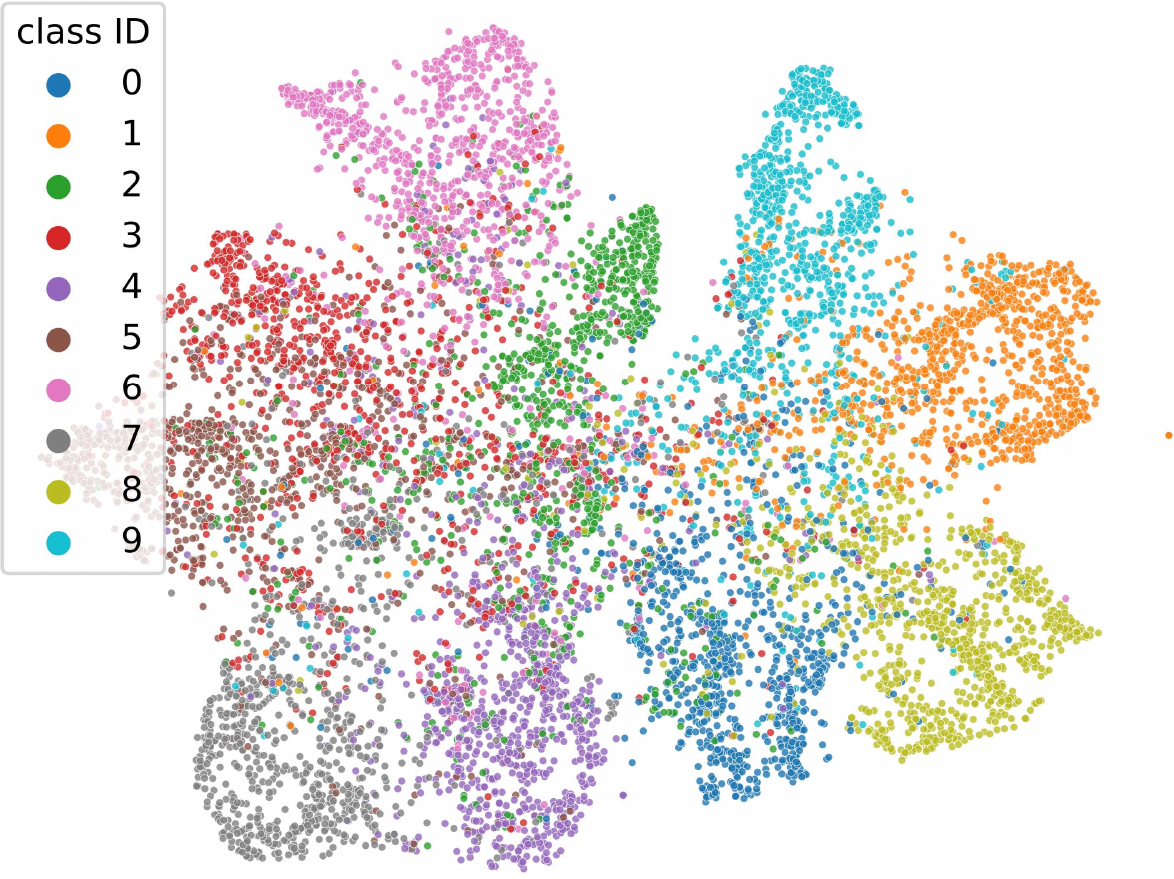}%
}
% \hfil
\subfloat[local model\label{fig:local}]{\includegraphics[width=0.5\linewidth]{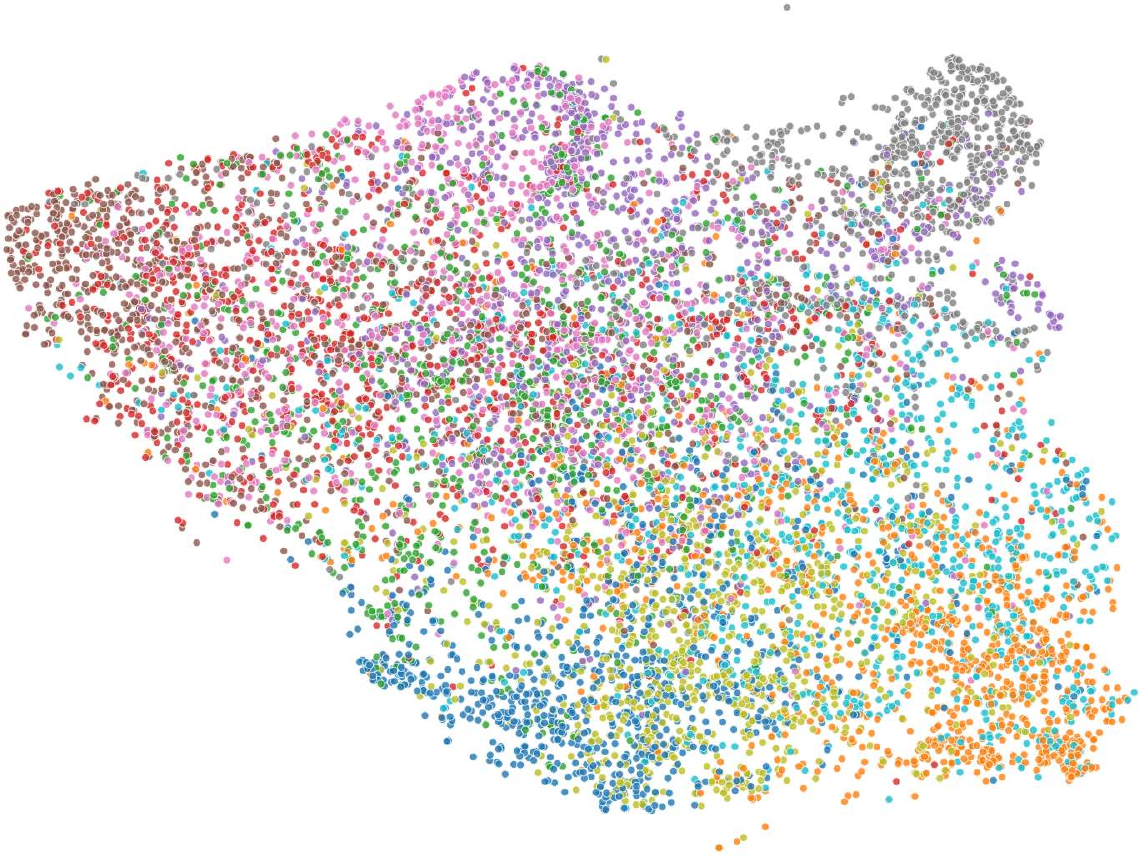}%
}
\hfil
\subfloat[FedAvg global model\label{fig:fedavg_global}]{\includegraphics[width=0.5\linewidth]{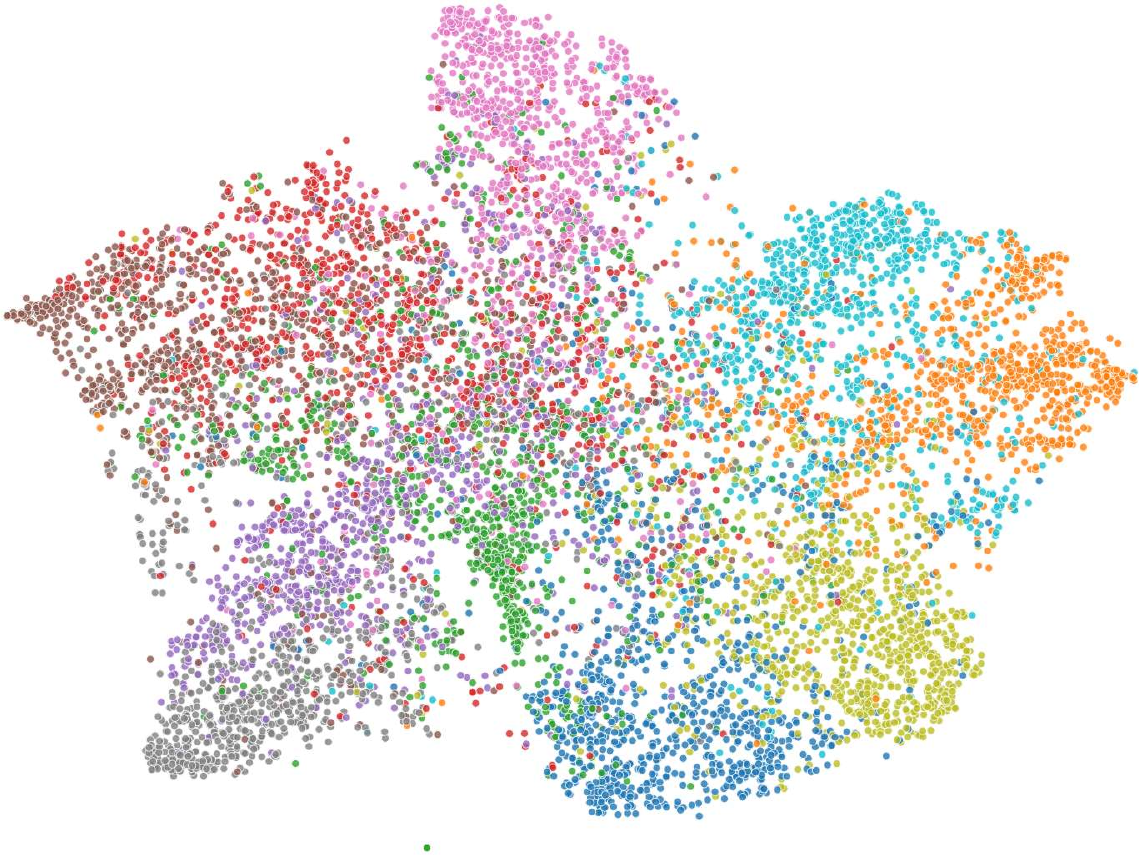}%
}
\subfloat[FedAvg local model\label{fig:fedavg_local}]{\includegraphics[width=0.5\linewidth]{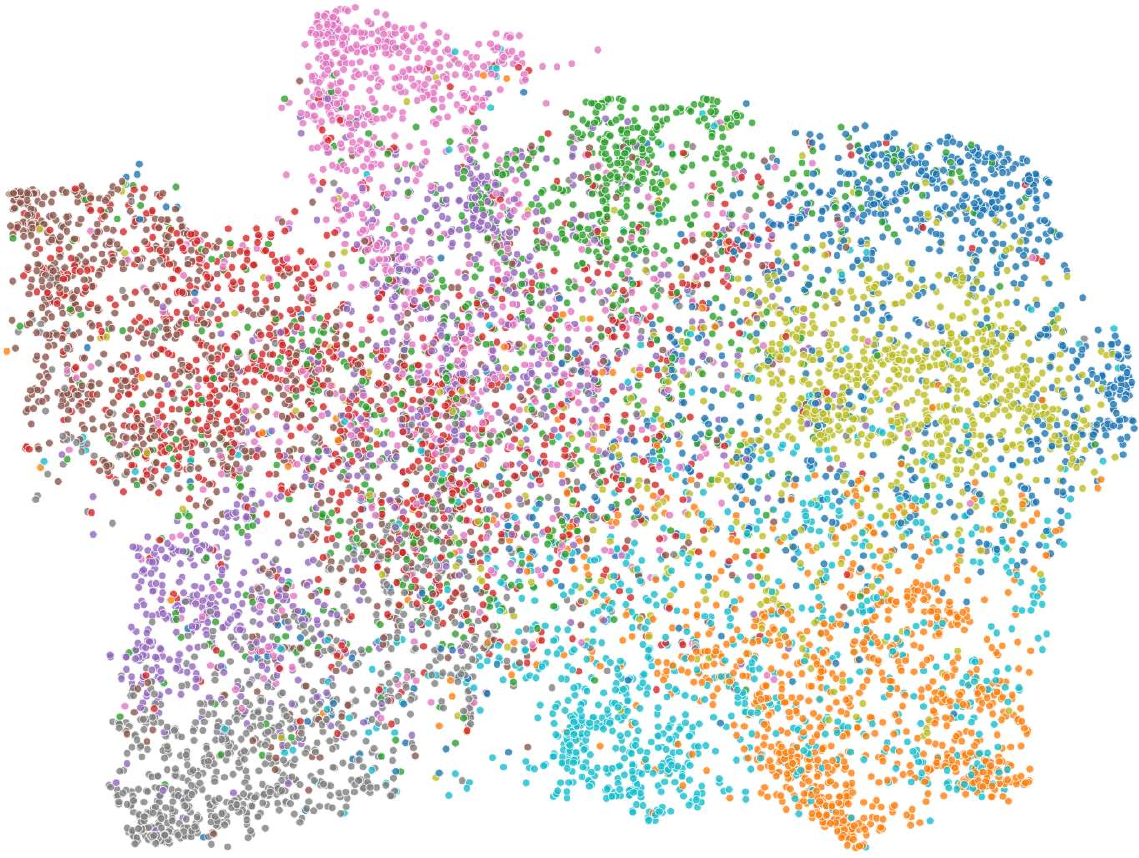}%
}
\caption{T-SNE visualizations of hidden vectors on CIFAR-10.}
\label{fig:nparty}
\vspace{-5pt}
\end{figure}

To further verify this intuition, we conduct a simple experiment on CIFAR-10. Specifically, we first train a CNN model (see Section \ref{sec:exp_setup} for the detailed structure) on CIFAR-10. We use t-SNE \cite{maaten2008visualizing} to visualize the hidden vectors of images from the test dataset as shown in Figure \ref{fig:global}. Then, we partition the dataset into 10 subsets in an unbalanced way (see Section \ref{sec:exp_setup} for the partition strategy) and train a CNN model on each subset. Figure \ref{fig:local} shows the t-SNE visualization of a randomly selected model. Apparently, the model trained on the subset learns poor features. The feature representations of most classes are even mixed and cannot be distinguished. Then, we run FedAvg algorithm on 10 subsets and show the representation learned by the global model in Figure \ref{fig:fedavg_global} and the representation learned by a selected local model (trained based on the global model) in Figure \ref{fig:fedavg_local}. We can observe that the points with the same class are more divergent in Figure \ref{fig:fedavg_local} compared with Figure \ref{fig:fedavg_global} (\eg, see class 9). The local training phase even leads the model to learn a worse representation due to the skewed local data distribution. This further verifies that the global model should be able to learn a better feature representation than the local model, and there is a \emph{drift} in the local updates. Therefore, under non-IID data scenarios, we should control the drift and bridge the gap between the representations learned by the local model and the global model.

\subsection{Method}
Based on the above intuition, we propose MOON. MOON is designed as a simple and effective approach based on FedAvg, only introducing lightweight but novel modifications in the local training phase. Since there is always drift in local training and the global model learns a better representation than the local model, MOON aims to decrease the distance between the representation learned by the local model and the representation learned by the global model, and increase the distance between the representation learned by the local model and the representation learned by the previous local model. We achieve this from the inspiration of contrastive learning, which is now mainly used to learn visual representations. In the following, we present the network architecture, the local learning objective and the learning procedure. At last, we discuss the relation to contrastive learning.

\subsubsection{Network Architecture}
The network has three components: a base encoder, a projection head, and an output layer. The base encoder is used to extract representation vectors from inputs. Like \cite{simclr}, an additional projection head is introduced to map the representation to a space with a fixed dimension. Last, as we study on the supervised setting, the output layer is used to produce predicted values for each class. For ease of presentation, with model weight $w$, we use $F_w(\cdot)$ to denote the whole network and $R_w(\cdot)$ to denote the network before the output layer (i.e., $R_w(x)$ is the mapped representation vector of input $x$).

\subsubsection{Local Objective}
\begin{figure}
    \centering
    \includegraphics[width=\linewidth]{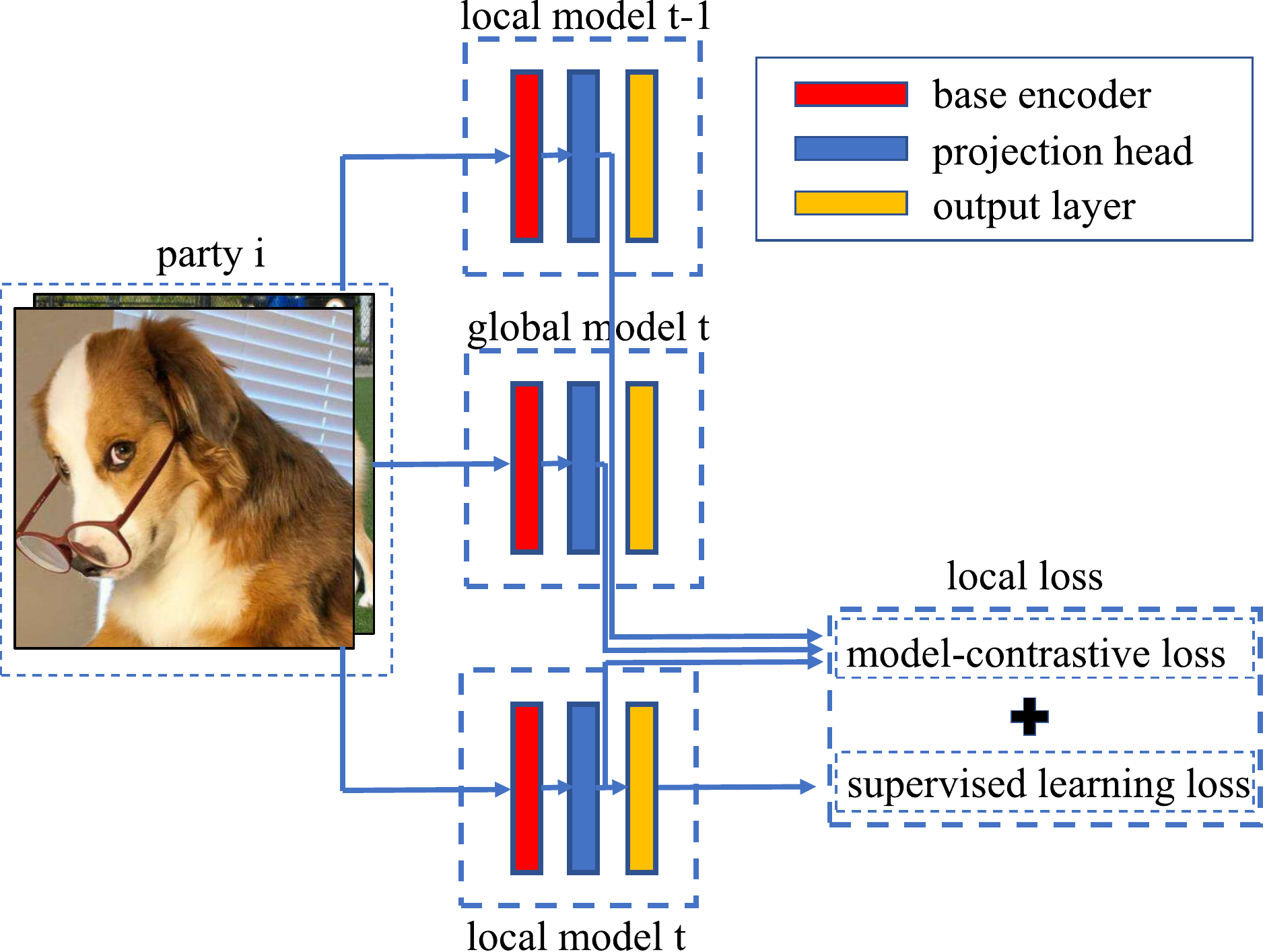}
    \caption{The local loss in MOON.}
    \label{fig:local_fram}
    % \vspace{-5pt}
\end{figure}

As shown in Figure \ref{fig:local_fram}, our local loss consists two parts. The first part is a typical loss term (\eg, cross-entropy loss) in supervised learning denoted as $\ell_{sup}$. The second part is our proposed model-contrastive loss term denoted as $\ell_{con}$.

Suppose party $P_i$ is conducting the local training. It receives the global model $w^t$ from the server and updates the model to $w_i^t$ in the local training phase. For every input $x$, we extract the representation of $x$ from the global model $w^t$ (i.e., $z_{glob} = R_{w^t}(x)$), the representation of $x$ from the local model of last round $w_i^{t-1}$ (i.e., $z_{prev} = R_{w_i^{t-1}}(x)$), and the representation of $x$ from the local model being updated $w_i^t$ (i.e., $z = R_{w_i^t}(x)$). Since the global model should be able to extract better representations, our goal is to decrease the distance between $z$ and $z_{glob}$, and increase the distance between $z$ and $z_{prev}$. Similar to NT-Xent loss \cite{sohn2016improved}, we define model-contrastive loss as

\begin{equation}
\begin{aligned}
\small
    &\ell_{con}  = -\log{\frac{\exp(\textrm{sim}(z, z_{glob})/\tau)}{\exp(\textrm{sim}(z, z_{glob})/\tau) + \exp(\textrm{sim}(z, z_{prev})/\tau)}} 
\end{aligned}
\end{equation}

where $\tau$ denotes a temperature parameter. The loss of an input $(x,y)$ is computed by 

\begin{equation}\label{eq:L}
    \ell = \ell_{sup}(w_i^t; (x,y)) + \mu\ell_{con}(w_i^t; w_i^{t-1}; w^t; x),
\end{equation}

where $\mu$ is a hyper-parameter to control the weight of model-contrastive loss. The local objective is to minimize 
\begin{equation}\label{eq:obj}
\min_{w_i^t} \mathbb{E}_{(x,y)\sim D^i} [\ell_{sup}(w_i^t; (x,y)) + \mu \ell_{con}(w_i^t; w_i^{t-1}; w^t; x)].
\end{equation}

The overall federated learning algorithm is shown in Algorithm \ref{alg:moon}. In each round, the server sends the global model to the parties, receives the local model from the parties, and updates the global model using weighted averaging. In local training, each party uses stochastic gradient descent to update the global model with its local data, while the objective is defined in Eq.~\eqref{eq:obj}.

\begin{algorithm}[t]
\SetNoFillComment
\LinesNumbered
\SetArgSty{textnormal}
\KwIn{number of communication rounds $T$, number of parties $N$, number of local epochs $E$, temperature $\tau$, learning rate $\eta$, hyper-parameter $\mu$}
\KwOut{The final model $w^T$}

\BlankLine
\tb{Server executes}:

initialize $w^0$\\
\For {$t=0, 1, ..., T-1$}{

    \For {$i=1, 2, ..., N$ \tb{in parallel}}{
    % \tcc{the server sends the global model to all parties}
        send the global model $w^t$ to $P_i$
        
        $w_i^t \leftarrow$ \tb{PartyLocalTraining}($i$, $w^t$)
    }
    $w^{t+1} \leftarrow \sum_{k=1}^N \frac{|\D^i|}{|\D|} w_k^{t}$
}

return $w^T$

\BlankLine
\tb{PartyLocalTraining}($i$, $w^t$):

$w_i^t \leftarrow w^t$ 

\For{epoch $i = 1, 2, ..., E$}{
    \For{each batch $\tb{b} = \{x, y\}$ of $\D^i$}{
        $\ell_{sup} \leftarrow CrossEntropyLoss(F_{w_i^t}(x), y)$
        
        $z \leftarrow R_{w_i^t}(x)$
        
        $z_{glob} \leftarrow R_{w^t}(x)$
        
        $z_{prev} \leftarrow R_{w_i^{t-1}}(x)$
        
        $\ell_{con} \leftarrow -\log{\frac{\exp(\textrm{sim}(z, z_{glob})/\tau)}{\exp(\textrm{sim}(z, z_{glob})/\tau) + \exp(\textrm{sim}(z, z_{prev})/\tau)}}$
        
        $\ell \leftarrow \ell_{sup} + \mu\ell_{con}$
        
        $w_i^t \leftarrow w_i^t - \eta \nabla \ell$
    }
}

return $w_i^t$ to server
\caption{The MOON framework}
\label{alg:moon}
\end{algorithm}

For simplicity, we describe MOON without applying sampling technique in Algorithm \ref{alg:moon}. MOON is still applicable when only a sample set of parties participate in federated learning each round. Like FedAvg, each party maintains its local model, which will be replaced
by the global model and updated only if the party is selected
to participate in a round. MOON only needs the latest local
model that the party has, even though it may not be updated
in round $(t-1)$ (e.g., $w_i^{t-1} = w_i^{t-2}$).

An notable thing is that considering an ideal case where the local model is good enough and learns (almost) the same representation as the global model (i.e., $z_{glob} = z_{prev}$), the model-contrastive loss will be a constant (i.e., $-\log\frac{1}{2}$). Thus, MOON will produce the same result as FedAvg, since there is no heterogeneity issue. In this sense, our approach is robust regardless of different amount of drifts.

\subsection{Comparisons with Contrastive Learning}
\begin{figure}
\centering
\includegraphics[width=\linewidth]{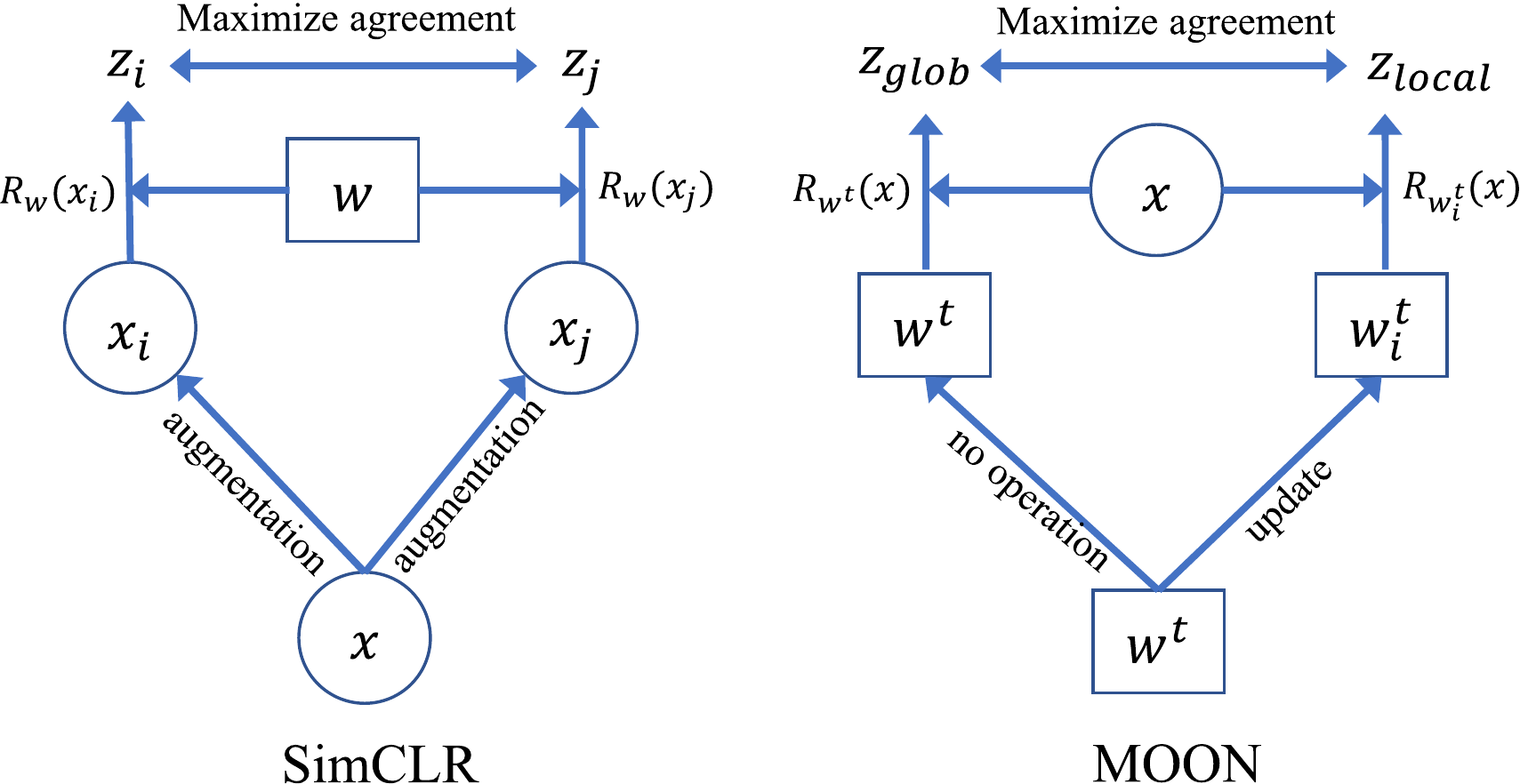}
\caption{The comparison between SimCLR and MOON. Here $x$ denotes an image, $w$ denotes a model, and $R$ denotes the function to compute representation. SimCLR maximizes the agreement between representations of different views of the same image, while MOON maximizes the agreement between representations of the local model and the global model {on the mini-batches.}}
\label{fig:comp}
% \vspace{-5pt}
\end{figure}

A comparison between MOON and SimCLR is shown in Figure \ref{fig:comp}. The model-contrastive loss compares representations learned by different models, while the contrastive loss compares representations of different images. We also highlight the key difference between MOON and traditional contrastive learning: MOON is currently for supervised learning in a federated setting while contrastive learning is for unsupervised learning in a centralized setting. Drawing the inspirations from contrastive learning, MOON is a new learning methodology in handling non-IID data distributions among parties in federated learning. 

\section{Experiments}
\label{sec:exp}
\subsection{Experimental Setup}
\label{sec:exp_setup}
We compare MOON with three state-of-the-art approaches including (1) FedAvg \cite{mcmahan2016communication},  (2) FedProx \cite{lifedprox}, and (3) SCAFFOLD \cite{karimireddy2019scaffold}.  We also compare a baseline approach named SOLO, where each party trains a model with its local data without federated learning. We conduct experiments on three datasets including CIFAR-10, CIFAR-100, and Tiny-Imagenet\footnote{\url{https://www.kaggle.com/c/tiny-imagenet}} (100,000 images with 200 classes). Moreover, we try two different network architectures. For CIFAR-10, we use a CNN network as the base encoder, which has two 5x5 convolution layers followed by 2x2 max pooling (the first with 6 channels and the second with 16 channels) and two fully connected layers with ReLU activation (the first with 120 units and the second with 84 units). For CIFAR-100 and Tiny-Imagenet, we use ResNet-50 \cite{he2016deep} as the base encoder. For all datasets, like \cite{simclr}, we use a 2-layer MLP as the projection head. The output dimension of the projection head is set to 256 by default. Note that all baselines use the same network architecture as MOON (including the projection head) for fair comparison.

We use PyTorch \cite{paszke2019pytorch} to implement MOON and the other baselines. The code is publicly available\footnote{\url{https://github.com/QinbinLi/MOON}}. We use the SGD optimizer with a learning rate 0.01 for all approaches. The SGD weight decay is set to 0.00001 and the SGD momentum is set to 0.9. The batch size is set to 64. The number of local epochs is set to 300 for SOLO. The number of local epochs is set to 10 for all federated learning approaches unless explicitly specified. The number of communication rounds is set to 100 for CIFAR-10/100 and 20 for Tiny-ImageNet, where all federated learning approaches have little or no accuracy gain with more communications. For MOON, we set the temperature parameter to 0.5 by default like \cite{simclr}.

\begin{figure*}[!]
\centering
\subfloat[CIFAR-10]{\includegraphics[width=0.32\textwidth]{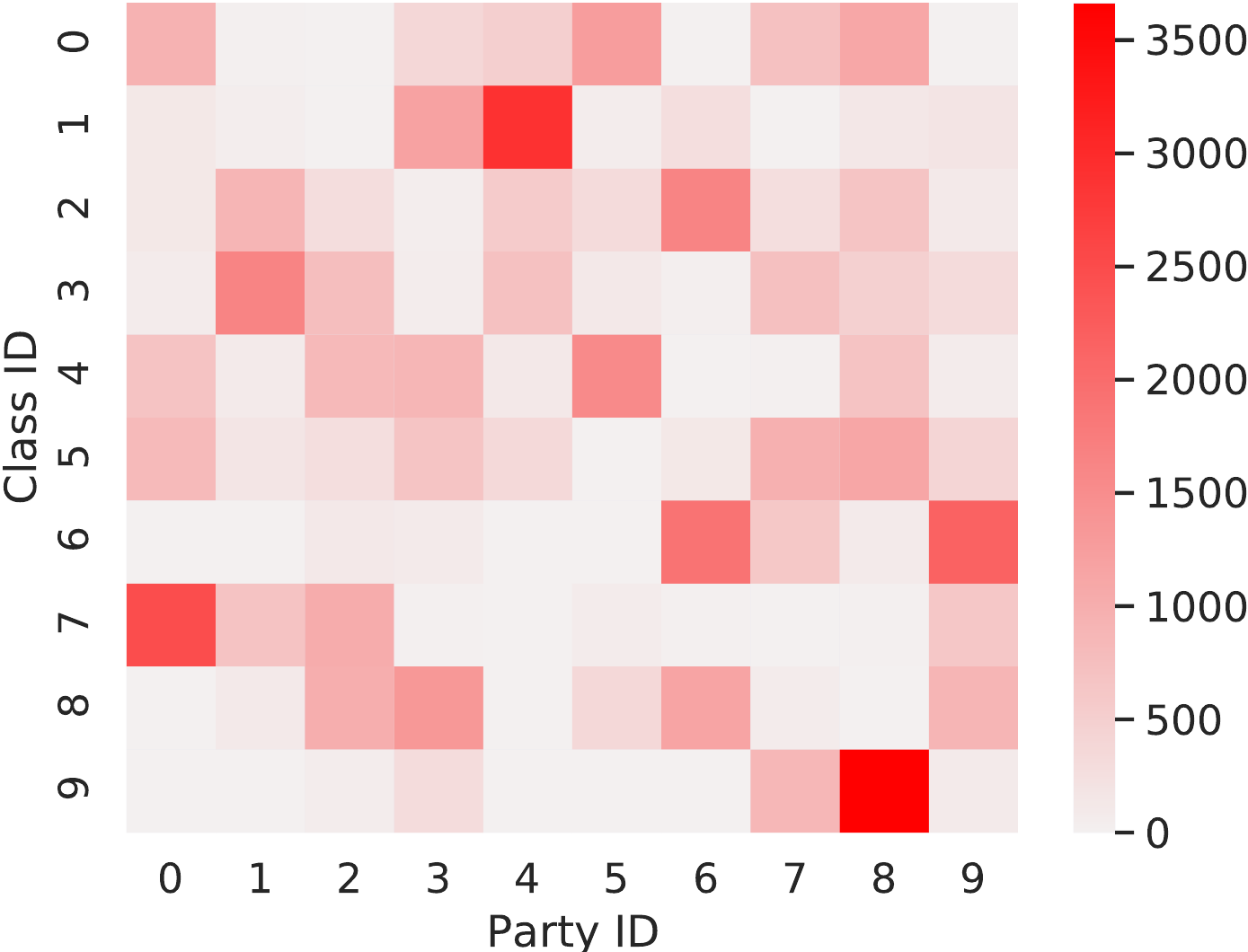}%
}
\hfill
% \hfil
\subfloat[CIFAR-100]{\includegraphics[width=0.32\textwidth]{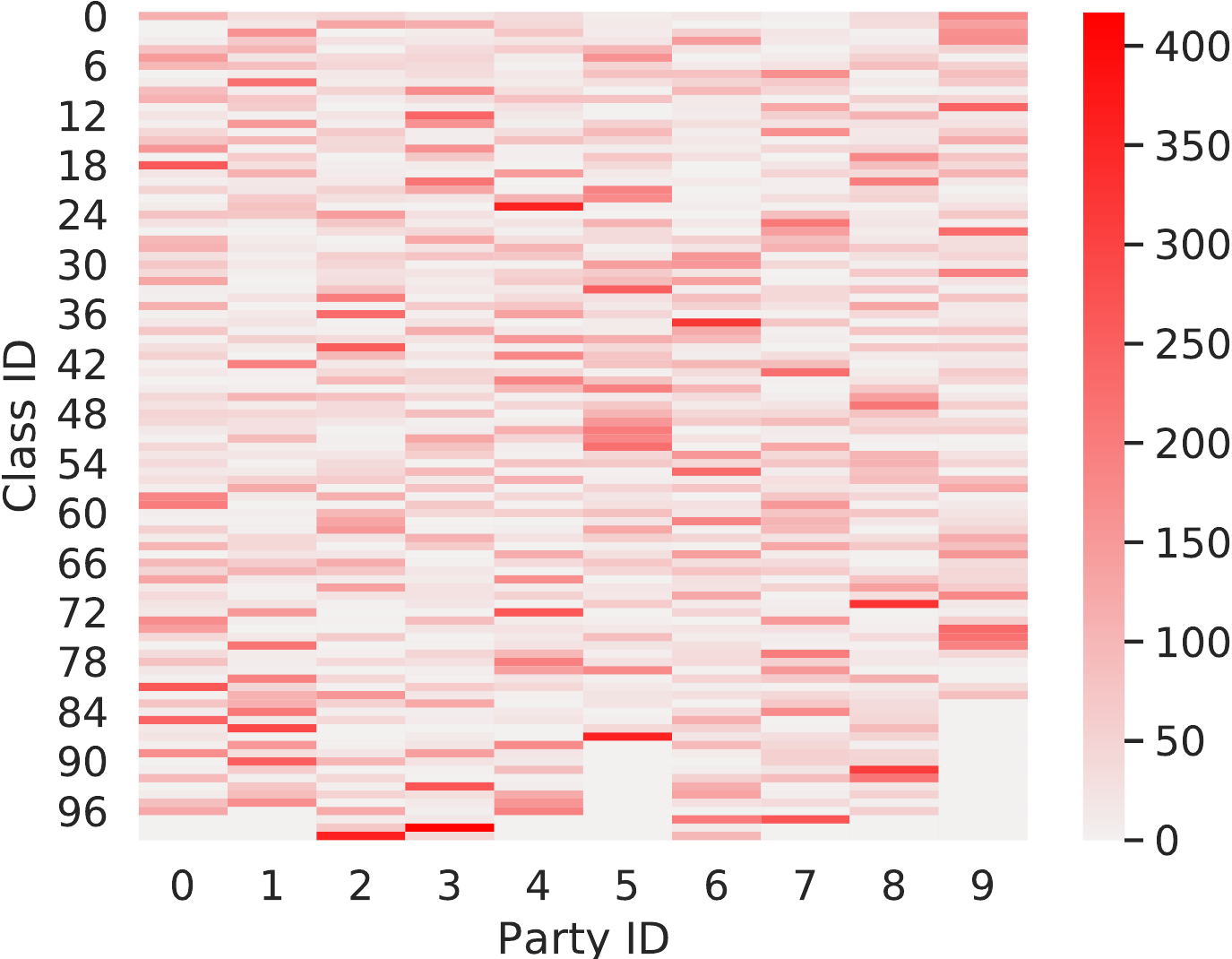}%
}
\hfill
\subfloat[Tiny-Imagenet]{\includegraphics[width=0.32\textwidth]{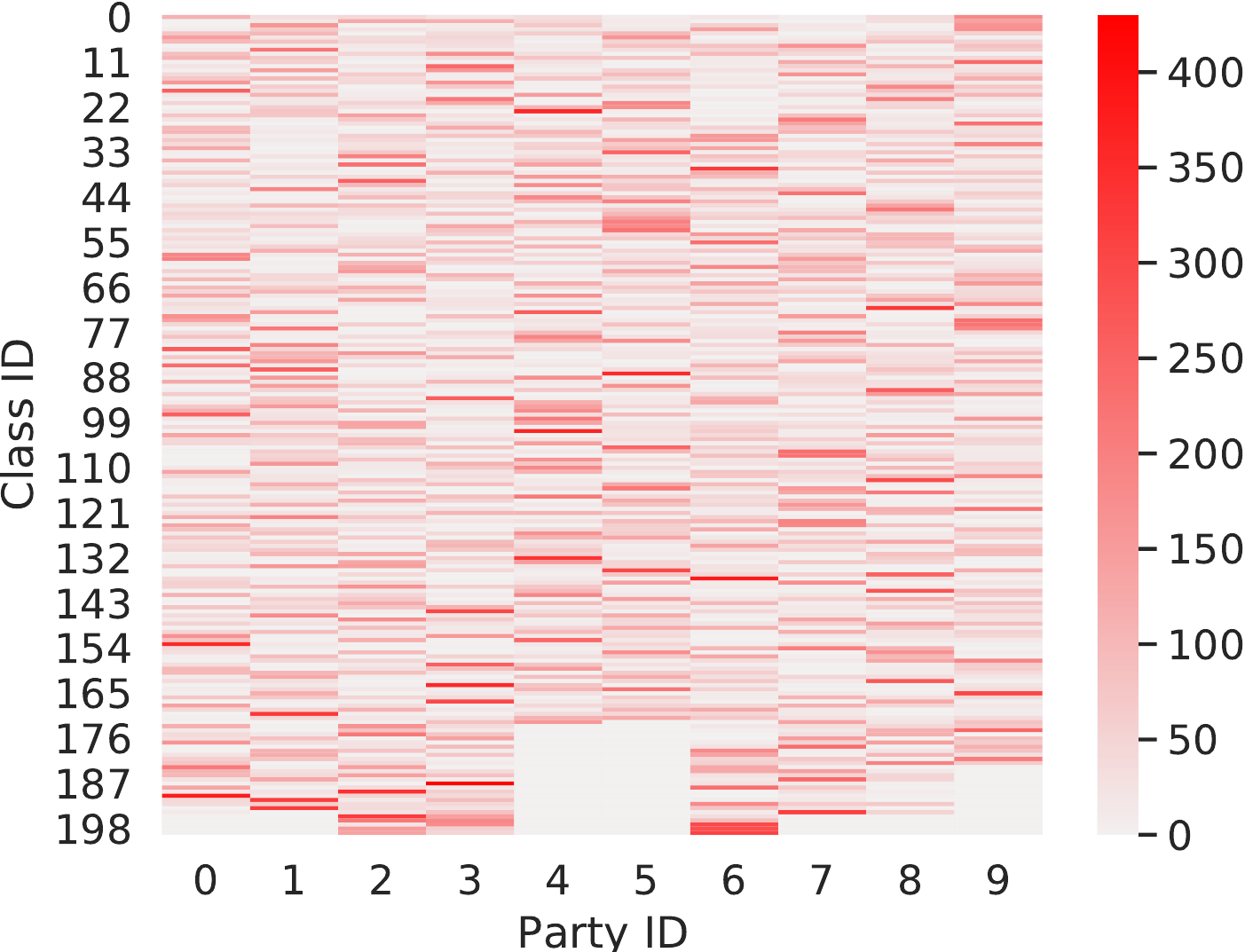}}
\caption{The data distribution of each party using non-IID data partition. The color bar denotes the number of data samples. Each rectangle represents the number of data samples of a specific class in a party. }
\label{fig:datadis}
\end{figure*}

Like previous studies \cite{pmlr-v97-yurochkin19a,Wang2020Federated}, we use Dirichlet distribution to generate the non-IID data partition among parties. Specifically, we sample $p_k \sim Dir_N(\beta)$ and allocate a $p_{k,j}$ proportion of the instances of class $k$ to party $j$, where $Dir(\beta)$ is the Dirichlet distribution with a concentration parameter $\beta$ (0.5 by default). With the above partitioning strategy, each party can have relatively few (even no) data samples in some classes. We set the number of parties to 10 by default. The data distributions among parties in default settings are shown in Figure \ref{fig:datadis}. For more experimental results, please refer to Appendix.

\subsection{Accuracy Comparison}
\label{sec:overall_acc}
For MOON, we tune $\mu$ from $\{0.1, 1, 5, 10\}$ and report the best result. The best $\mu$ of MOON for CIFAR-10, CIFAR-100, and Tiny-Imagenet are 5, 1, and 1, respectively. Note that FedProx also has a hyper-parameter $\mu$ to control the weight of its proximal term (i.e., $L_{FedProx} = \ell_{FedAvg} + \mu \ell_{prox}$). For FedProx, we tune $\mu$ from $\{0.001, 0.01, 0.1, 1\}$ (the range is also used in the previous paper \cite{lifedprox}) and report the best result. The best $\mu$ of FedProx for CIFAR-10, CIFAR-100, and Tiny-Imagenet are $0.01$, $0.001$, and $0.001$, respectively. Unless explicitly specified, we use these $\mu$ settings for all the remaining experiments.

Table \ref{tbl:allacc} shows the top-1 test accuracy of all approaches with the above default setting. Under non-IID settings, SOLO demonstrates much worse accuracy than other federated learning approaches. This demonstrates the benefits of federated learning. Comparing different federated learning approaches, we can observe that MOON is consistently the best approach among all tasks. It can outperform FedAvg by 2.6\% accuracy on average of all tasks. For FedProx, its accuracy is very close to FedAvg. The proximal term in FedProx has little influence in the training since $\mu$ is small. However, when $\mu$ is not set to a very small value, the convergence of FedProx is quite slow (see Section \ref{sec:comm_effi}) and the accuracy of FedProx is bad. For SCAFFOLD, it has much worse accuracy on CIFAR-100 and Tiny-Imagenet than other federated learning approaches.

\begin{table}[]
\centering
\caption{The top-1 accuracy of MOON and the other baselines on test datasets. For MOON, FedAvg, FedProx, and SCAFFOLD, we run three trials and report the mean and standard derivation. For SOLO, we report the mean and standard derivation among all parties.}
\label{tbl:allacc}
\resizebox{\linewidth}{!}{%
\begin{tabular}{|c|c|c|c|}
\hline
Method & CIFAR-10 & CIFAR-100 & Tiny-Imagenet \\ \hline \hline
MOON & \tb{69.1\%}$\pm$0.4\% & \tb{67.5\%}$\pm$0.4\% & \tb{25.1\%}$\pm$0.1\%  \\ \hline
FedAvg & 66.3\%$\pm$0.5\% & 64.5\% $\pm$0.4\% & 23.0\%$\pm$0.1\%   \\ \hline
FedProx & 66.9\%$\pm$0.2\% & 64.6\%$\pm$0.2\% & 23.2\%$\pm$0.2\% \\ \hline
SCAFFOLD & 66.6\%$\pm$0.2\% & 52.5\% $\pm$0.3\% & 16.0\%$\pm$0.2\%  \\ \hline
SOLO &46.3\% $\pm$5.1\% &22.3\%$\pm$1.0\% &8.6\%$\pm$0.4\%\\\hline
\end{tabular}
}
\end{table}

\subsection{Communication Efficiency}
\label{sec:comm_effi}
\begin{figure*}[!]
\centering
\subfloat[CIFAR-10]{\includegraphics[width=0.32\textwidth]{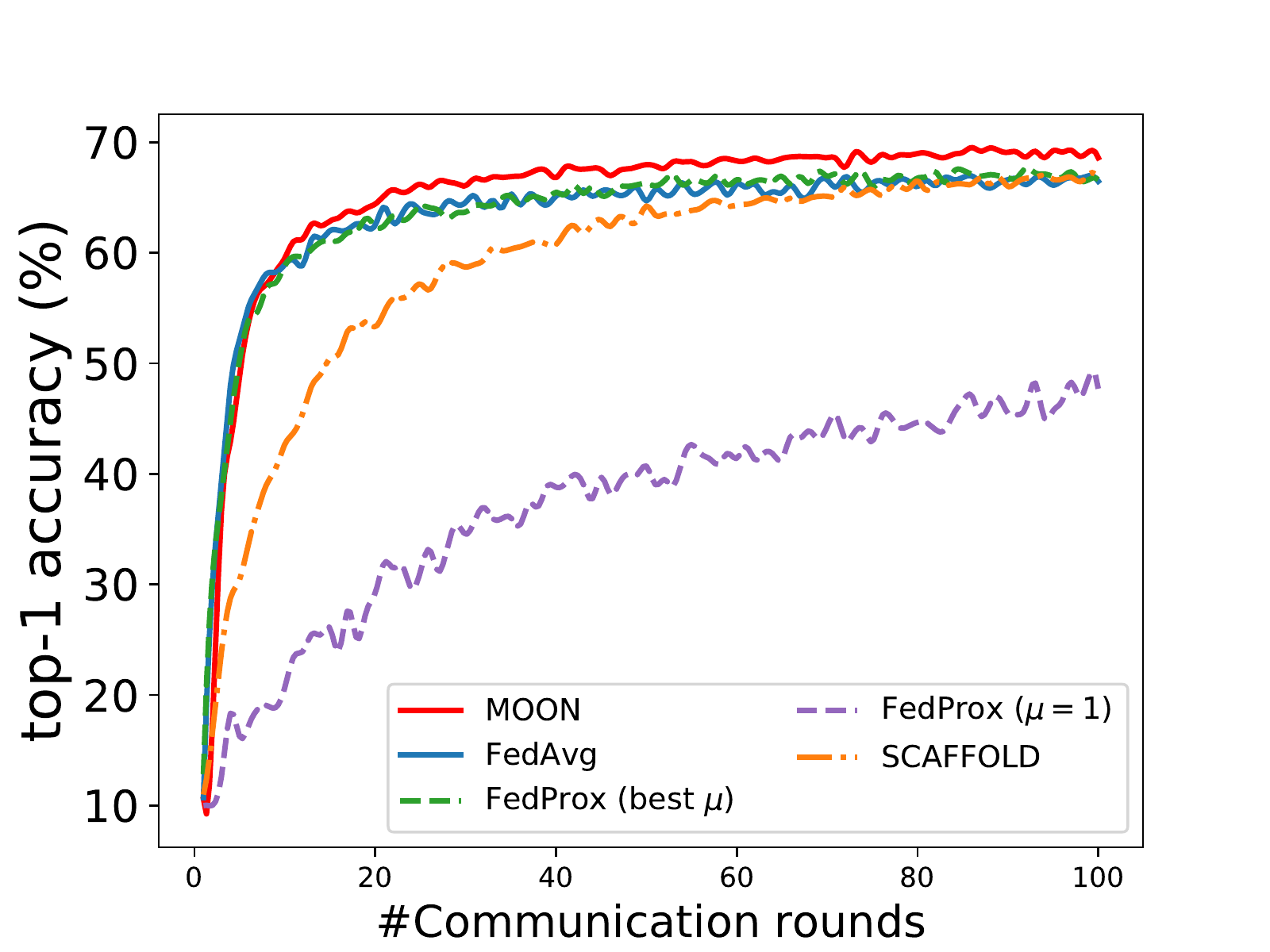}%
}
\hfill
% \hfil
\subfloat[CIFAR-100]{\includegraphics[width=0.32\textwidth]{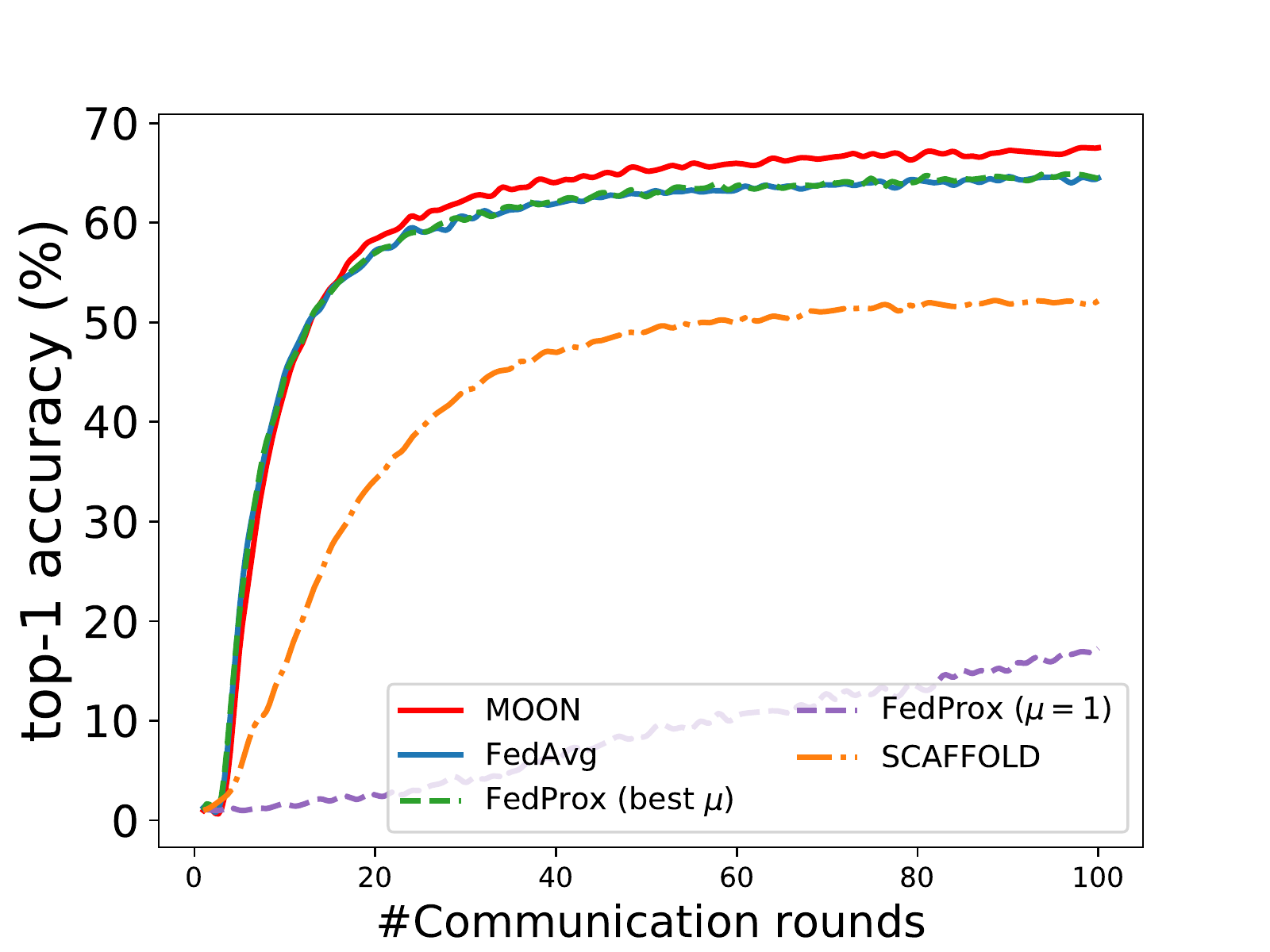}%
}
\hfill
\subfloat[Tiny-Imagenet]{\includegraphics[width=0.32\textwidth]{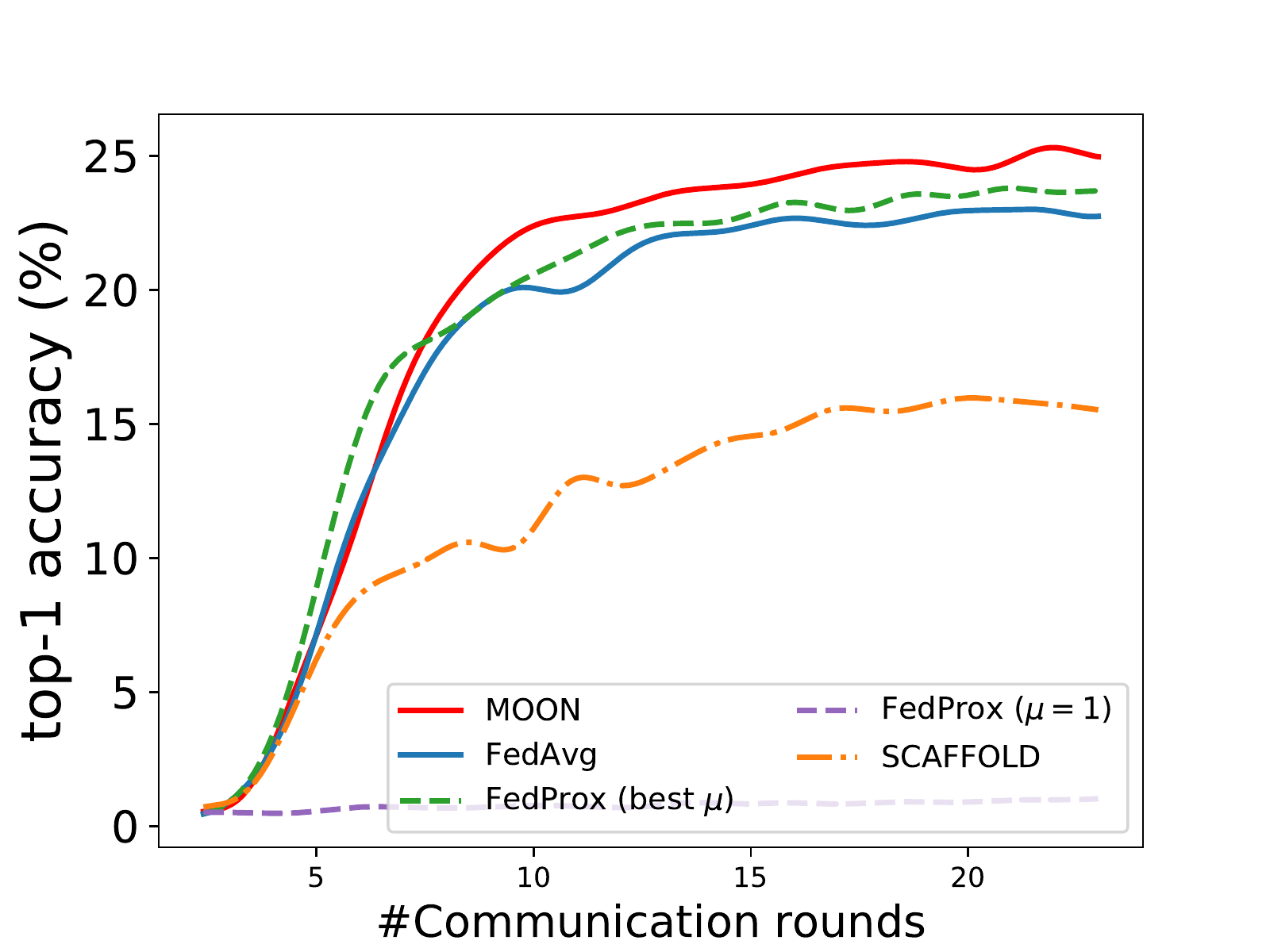}}
\caption{The top-1 test accuracy in different number of communication rounds. For FedProx, we report both the accuracy with best $\mu$ and the accuracy with $\mu=1$.}
\label{fig:comm}
\vspace{-15pt}
\end{figure*}

Figure \ref{fig:comm} shows the accuracy in each round during training. As we can see, the model-contrastive loss term has little influence on the convergence rate with best $\mu$. The speed of accuracy improvement in MOON is almost the same as FedAvg at the beginning, while it can achieve a better accuracy later benefit from the model-contrastive loss. Since the best $\mu$ values are generally small in FedProx, FedProx with best $\mu$ is very close to FedAvg, especially on CIFAR-10 and CIFAR-100. However, when setting $\mu=1$, FedProx becomes very slow due to the additional proximal term. This implies that limiting the $\ell_2$-norm distance between the local model and the global model is not an effective solution. Our model-contrastive loss can effectively increase the accuracy without slowing down the convergence.

We show the number of communication rounds to achieve the same accuracy as running FedAvg for 100 rounds on CIFAR-10/100 or 20 rounds on Tiny-Imagenet in Table \ref{tbl:comm}. We can observe that the number of communication rounds is significantly reduced in MOON. MOON needs about half the number of communication rounds on CIFAR-100 and Tiny-Imagenet compared with FedAvg. On CIFAR-10, the speedup of MOON is even close to 4. MOON is much more communication-efficient than the other approaches.

\begin{table}[]
\centering
\caption{The number of rounds of different approaches to achieve the same accuracy as running FedAvg for 100 rounds (CIFAR-10/100) or 20 rounds (Tiny-Imagenet). The speedup of an approach is computed against FedAvg.}
\label{tbl:comm}
\resizebox{\linewidth}{!}{%
\begin{tabular}{|c|c|c|c|c|c|c|}
\hline
 \multirow{2}{*}{Method} & \multicolumn{2}{c|}{CIFAR-10} & \multicolumn{2}{c|}{CIFAR-100} & \multicolumn{2}{c|}{Tiny-Imagenet} \\ \cline{2-7} 
 & \#rounds & speedup & \#rounds & speedup & \#rounds & speedup \\ \hline \hline
FedAvg & 100 & 1$\times$ & 100 & 1$\times$ &20  & 1$\times$ \\ \hline
FedProx & 52 & 1.9$\times$ & 75 & 1.3$\times$ & 17 & 1.2$\times$ \\ \hline
SCAFFOLD & 80 & 1.3$\times$ & \backslashbox{}{} & \textless{}1$\times$  & \backslashbox{}{} & \textless{}1$\times$  \\ \hline
MOON & \tb{27} & \tb{3.7}$\times$ & \tb{43} & \tb{2.3}$\times$ & \tb{11} & \tb{1.8}$\times$ \\ \hline
\end{tabular}
}
\end{table}

\begin{figure*}
\centering
\subfloat[CIFAR-10]{\includegraphics[width=0.32\textwidth]{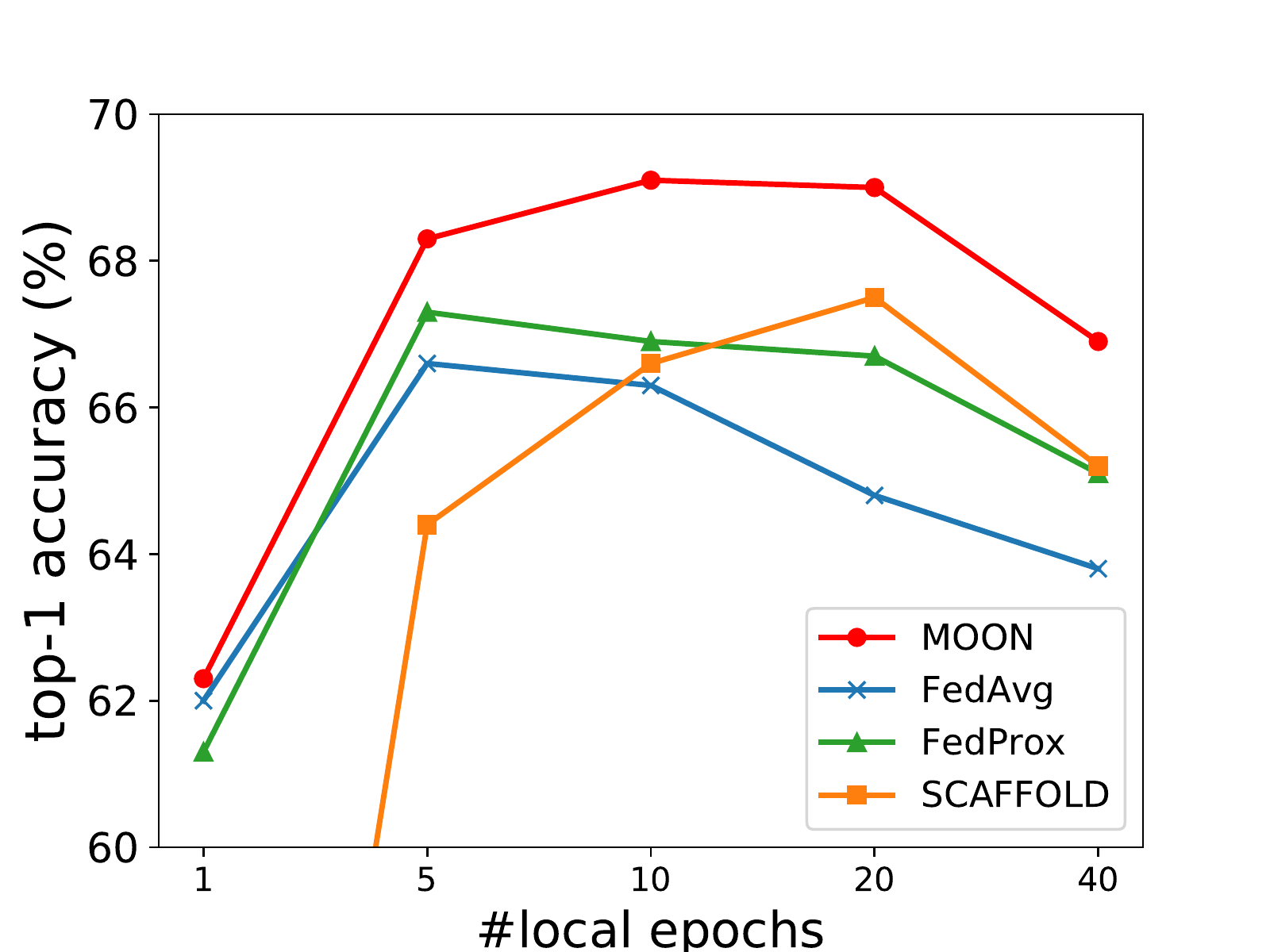}%
}
\hfill
% \hfil
\subfloat[CIFAR-100]{\includegraphics[width=0.32\textwidth]{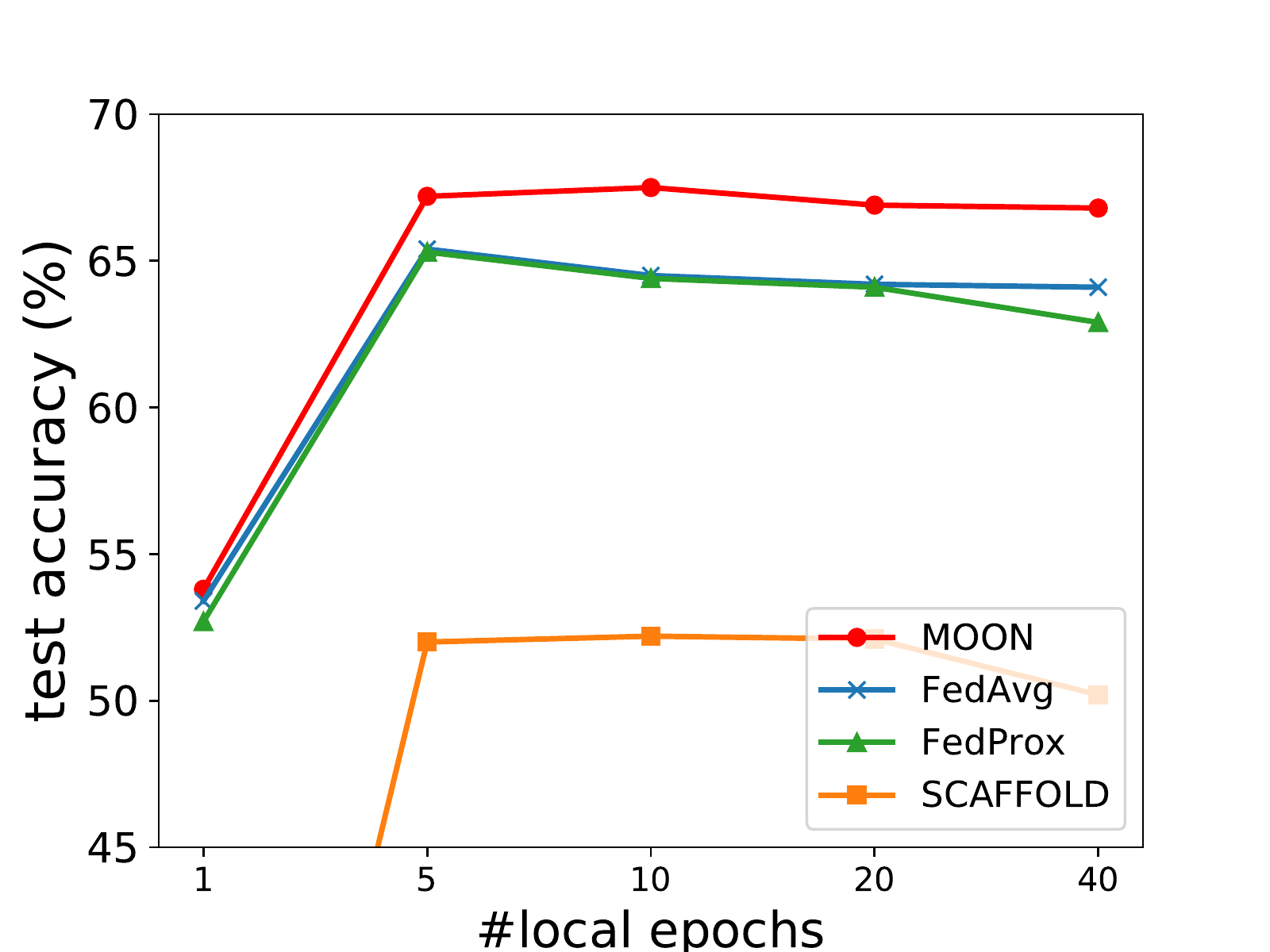}%
}
\hfill
\subfloat[Tiny-Imagenet]{\includegraphics[width=0.32\textwidth]{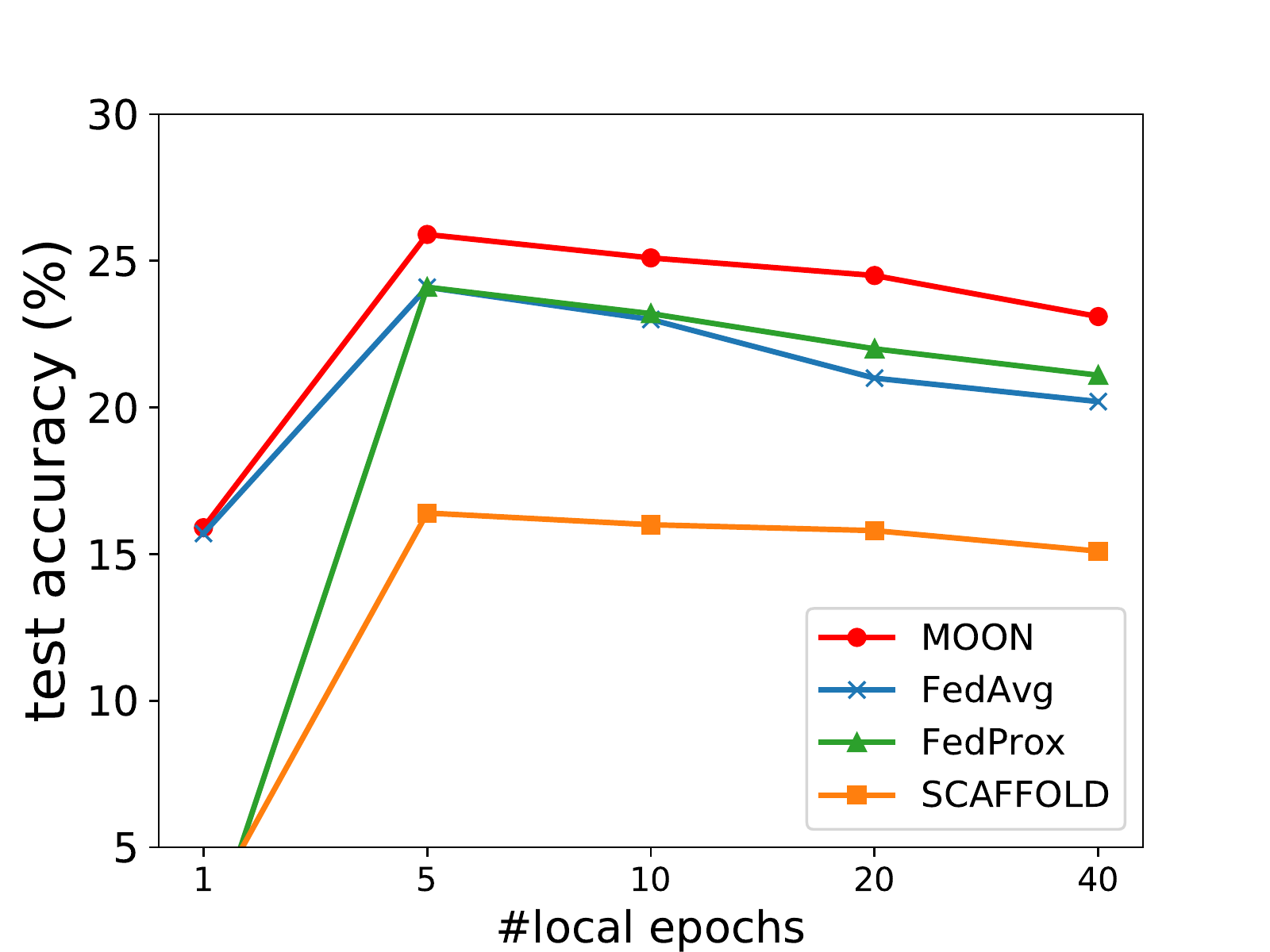}}
\caption{The top-1 test accuracy with different number of local epochs. For MOON and FedProx, $\mu$ is set to the best $\mu$ from Section \ref{sec:overall_acc} for all numbers of local epochs. The accuracy of SCAFFOLD is quite bad when number of local epochs is set to 1 (45.3\% on CIFAR10, 20.4\% on CIFAR-100, 2.6\% on Tiny-Imagenet). The accuracy of FedProx on Tiny-Imagenet with one local epoch is 1.2\%.}
\label{fig:epoch}
\vspace{-5pt}
\end{figure*}

\subsection{Number of Local Epochs}
We study the effect of number of local epochs on the accuracy of final model. The results are shown in Figure \ref{fig:epoch}. When the number of local epochs is 1, the local update is very small. Thus, the training is slow and the accuracy is relatively low given the same number of communication rounds. All approaches have a close accuracy (MOON is still the best). When the number of local epochs becomes too large, the accuracy of all approaches drops, which is due to the drift of local updates, i.e., the local optima are not consistent with the global optima. Nevertheless, MOON clearly outperforms the other approaches. This further verifies that MOON can effectively mitigate the negative effects of the drift by too many local updates.

\subsection{Scalability}

\begin{figure*}
\centering
\subfloat[50 parties]{\includegraphics[width=0.45\linewidth]{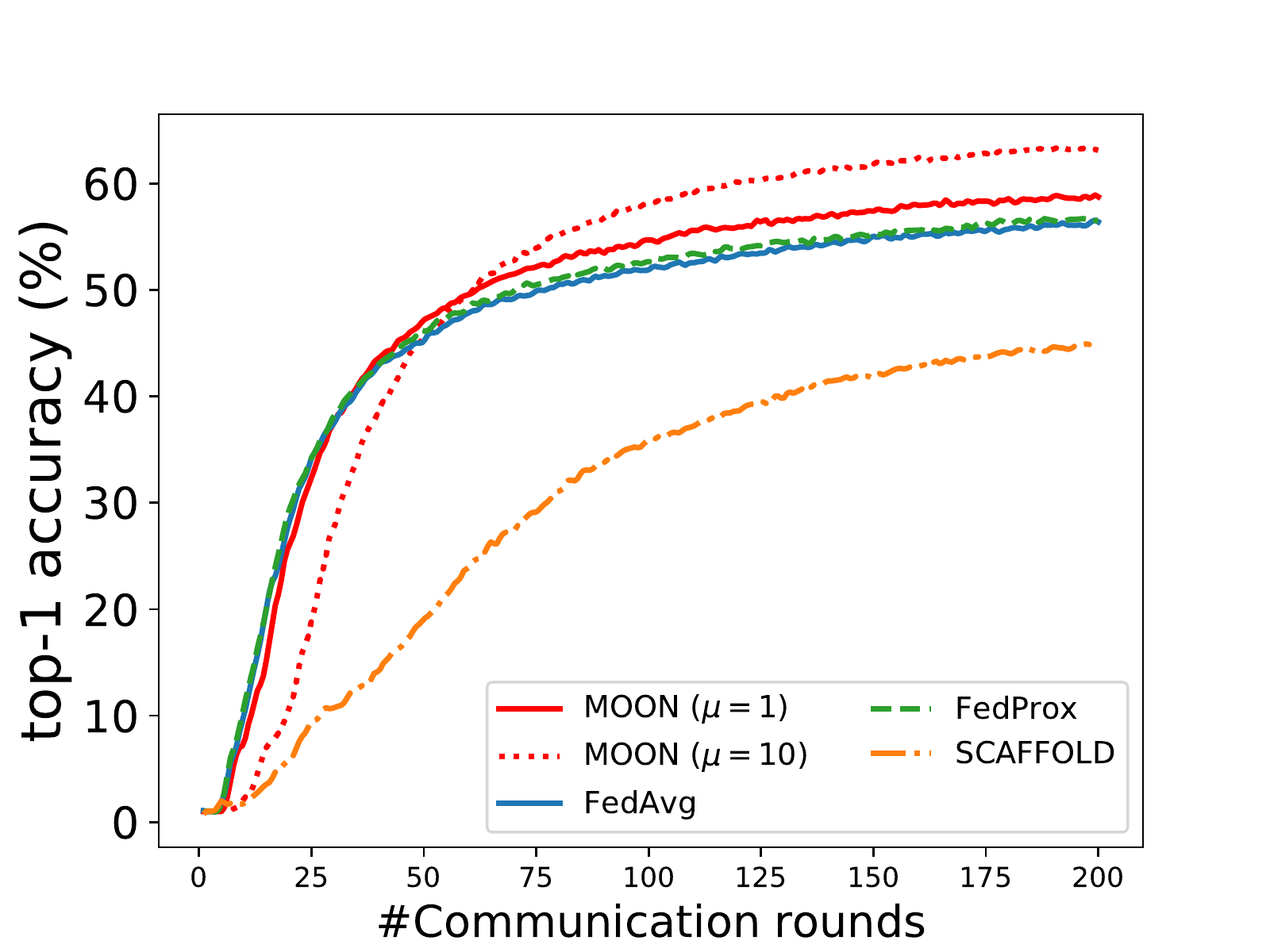}
}
\hfil
\subfloat[100 parties (sample fraction=0.2)]{\includegraphics[width=0.45\linewidth]{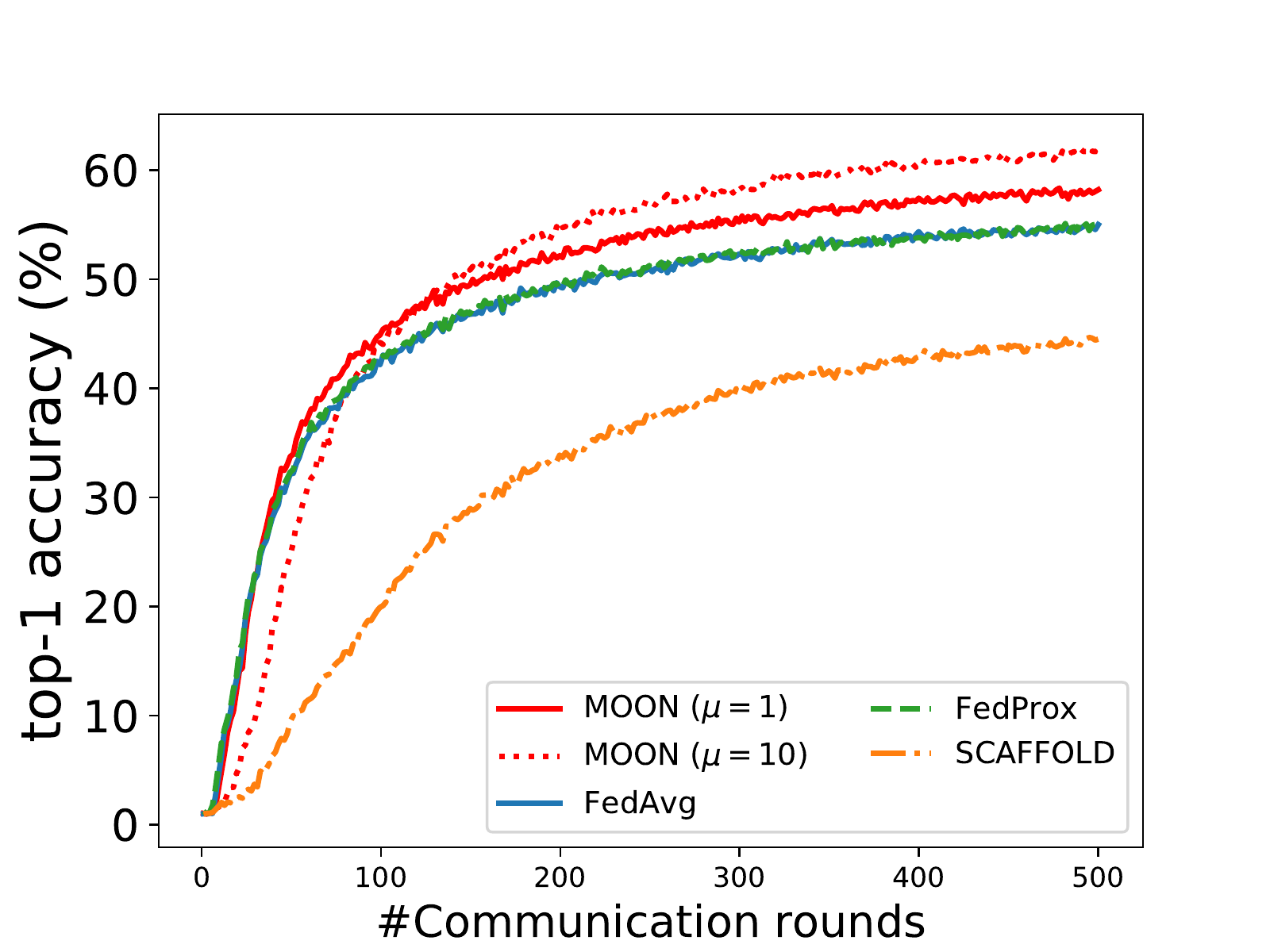}%
}
\hfil
\caption{The top-1 test accuracy on CIFAR-100 with 50/100 parties.}
\label{fig:scala}
\end{figure*}

To show the scalability of MOON, we try a larger number of parties on CIFAR-100. Specifically, we try two settings: (1) We partition the dataset into 50 parties and all parties participate in federated learning in each round. (2) We partition the dataset into 100 parties and randomly sample 20 parties to participate in federated learning in each round (client sampling technique introduced in FedAvg \cite{mcmahan2016communication}). The results are shown in Table \ref{tbl:scala} and Figure \ref{fig:scala}. For MOON, we show the results with $\mu=1$ (best $\mu$ from Section \ref{sec:overall_acc}) and $\mu=10$. For MOON ($\mu=1$), it outperforms the FedAvg and FedProx over 2\% accuracy at 200 rounds with 50 parties and 3\% accuracy at 500 rounds with 100 parties. Moreover, for MOON ($\mu=10$), although the large model-contrastive loss slows down the training at the beginning as shown in Figure \ref{fig:scala}, MOON can outperform the other approaches a lot with more communication rounds. Compared with FedAvg and FedProx, MOON achieves about about 7\% higher accuracy at 200 rounds with 50 parties and at 500 rounds with 100 parties. SCAFFOLD has a low accuracy with a relatively large number of parties.

\begin{table}[]
\centering
\caption{The accuracy with 50 parties and 100 parties (sample fraction=0.2) on CIFAR-100.}
\label{tbl:scala}
\resizebox{\linewidth}{!}{%
\begin{tabular}{|c|c|c|c|c|}
\hline
\multirow{2}{*}{Method} & \multicolumn{2}{c|}{\#parties=50} & \multicolumn{2}{c|}{\#parties=100} \\ \cline{2-5} 
  & 100 rounds & 200 rounds & 250 rounds & 500 rounds \\ \hline \hline
MOON ($\mu$=1) & 54.7\% &58.8\% & 54.5\% & 58.2\% \\ \hline
MOON ($\mu$=10)& \tb{58.2\%} &\tb{63.2\%} & \tb{56.9\%} & \tb{61.8\%} \\ \hline
FedAvg & 51.9\% &56.4\% & 51.0\% & 55.0\% \\ \hline
FedProx  & 52.7\% & 56.6\% & 51.3\% & 54.6\% \\ \hline
SCAFFOLD & 35.8\% & 44.9\% & 37.4\% & 44.5\% \\ \hline
SOLO & \multicolumn{2}{c|}{10\%$\pm$0.9\%} & \multicolumn{2}{c|}{7.3\%$\pm$0.6\%} \\ \hline
\end{tabular}
}
\end{table}

\subsection{Heterogeneity}
We study the effect of data heterogeneity by varying the concentration parameter $\beta$ of Dirichlet distribution on CIFAR-100. For a smaller $\beta$, the partition will be more unbalanced. The results are shown in Table \ref{tbl:beta}. MOON always achieves the best accuracy among three unbalanced levels. When the unbalanced level decreases (i.e., $\beta=5$), FedProx is worse than FedAvg, while MOON still outperforms FedAvg with more than 2\% accuracy. The experiments demonstrate the effectiveness and robustness of MOON.

\begin{table}[]
\centering
\caption{The test accuracy with $\beta$ from \{0.1, 0.5, 5\}. }
\label{tbl:beta}
\resizebox{\linewidth}{!}{%
\begin{tabular}{|c|c|c|c|}
\hline
Method & $\beta=0.1$ & $\beta=0.5$ & $\beta=5$ \\ \hline \hline
MOON & \tb{64.0\%} & \tb{67.5\%} & \tb{68.0\%} \\ \hline
FedAvg & 62.5\% & 64.5\% & 65.7\% \\ \hline
FedProx & 62.9\% & 64.6\% & 64.9\% \\ \hline
SCAFFOLD & 47.3\% & 52.5\% & 55.0\% \\ \hline
SOLO & 15.9\%$\pm$1.5\% & 22.3\%$\pm$1\% & 26.6\%$\pm$1.4\% \\ \hline
\end{tabular}
}
\end{table}

\subsection{Loss Function}

% \paragraph{Loss Function}
To maximize the agreement between the representation learned by the global model and the representation learned by the local model, our model-contrastive loss $\ell_{con}$ is proposed inspired by NT-Xent loss \cite{simclr}. Another intuitive option is to use $\ell_2$ regularization, and the local loss is
\begin{equation}
    \ell = \ell_{sup} + \mu \norm{z-z_{glob}}_2
\end{equation}

Here we compare the approaches using different kinds of loss functions to limit the representation: no additional term (i.e., FedAvg: $L = \ell_{sup}$), $\ell_2$ norm, and our model-contrastive loss. The results are shown in Table \ref{tbl:loss}. We can observe that simply using $\ell_2$ norm even cannot improve the accuracy compared with FedAvg on CIFAR-10. While using $\ell_2$ norm can improve the accuracy on CIFAR-100 and Tiny-Imagenet, the accuracy is still lower than MOON. Our model-contrastive loss is an effective way to constrain the representations. 

Our model-contrastive loss influences the local model from two aspects. Firstly, the local model learns a close representation to the global model. Secondly, the local model also learns a better representation than the previous one until the local model is good enough (i.e., $z=z_{glob}$ and $\ell_{con}$ becomes a constant).

\begin{table}[]
\centering
\caption{The top-1 accuracy with different kinds of loss for the second term of local objective. We tune $\mu$ from \{0.001, 0.01 , 0.1, 1, 5, 10\} for the $\ell_2$ norm approach and report the best accuracy.}
\label{tbl:loss}
\resizebox{\linewidth}{!}{
\begin{tabular}{|c|c|c|c|}
\hline
second term & CIFAR-10 & CIFAR-100 & Tiny-Imagenet \\ \hline\hline
none (FedAvg) & 66.3\% & 64.5\% & 23.0\% \\ \hline
$\ell_2$ norm & 65.8\% & 66.9\% & 24.0\% \\ \hline
MOON & \tb{69.1\%} & \tb{67.5\%} & \tb{25.1\%} \\ \hline
\end{tabular}
}
\end{table}

\section{Conclusion}
Federated learning has become a promising approach to resolve the pain of data silos in many domains such as medical imaging, object detection, and landmark classification. Non-IID is a key challenge for the effectiveness of federated learning. To improve the performance of federated deep learning models on non-IID datasets, we propose model-contrastive learning (MOON), a simple and effective approach for federated learning. MOON introduces a new learning concept, i.e., contrastive learning in model-level. Our extensive experiments show that MOON achieves significant improvement over state-of-the-art approaches on various image classification tasks. As MOON does not require the inputs to be images, it potentially can be applied to non-vision problems.

\section*{Acknowledgements}
This research is supported by the National Research Foundation, Singapore under its AI Singapore Programme (AISG Award No: AISG2-RP-2020-018). Any opinions, findings and conclusions or recommendations expressed in this material are those of the authors and do not reflect the views of National Research Foundation, Singapore. The authors thank Jianxin Wu, Chaoyang He, Shixuan Sun, Yaqi Xie and Yuhang Chen for their feedback. The authors also thank Yuzhi Zhao, Wei Wang, and Mo Sha for their supports of computing resources.

{\small
\bibliographystyle{ieee_fullname}
\bibliography{ref}
}

\appendix
% \newpage
\section{More Details of the Datasets}
The statistics of each dataset are shown in Table \ref{tbl:data} when setting $\beta=0.5$. All datasets provide a training dataset and test dataset. All the reported accuracies are computed on the test dataset.

\begin{table}[h]
\centering
\caption{The statistics of datasets.}
\label{tbl:data}
\resizebox{\linewidth}{!}{%
\begin{tabular}{|c|c|c|c|}
\hline
\multirow{2}{*}{dataset} & \multicolumn{2}{c|}{\#training samples/party} & \multirow{2}{*}{\#test samples} \\ \cline{2-3}
 & mean & std &  \\ \hline \hline
CIFAR10 & 5,000 & 1,165 & 10,000 \\ \hline
CIFAR100 & 5,000 & 181 & 10,000 \\ \hline
Tiny-Imagenet & 10,000 & 99 & 10,000 \\ \hline
\end{tabular}
}
\end{table}

\begin{figure*}
\centering
\subfloat[CIFAR-10]{\includegraphics[width=0.32\textwidth]{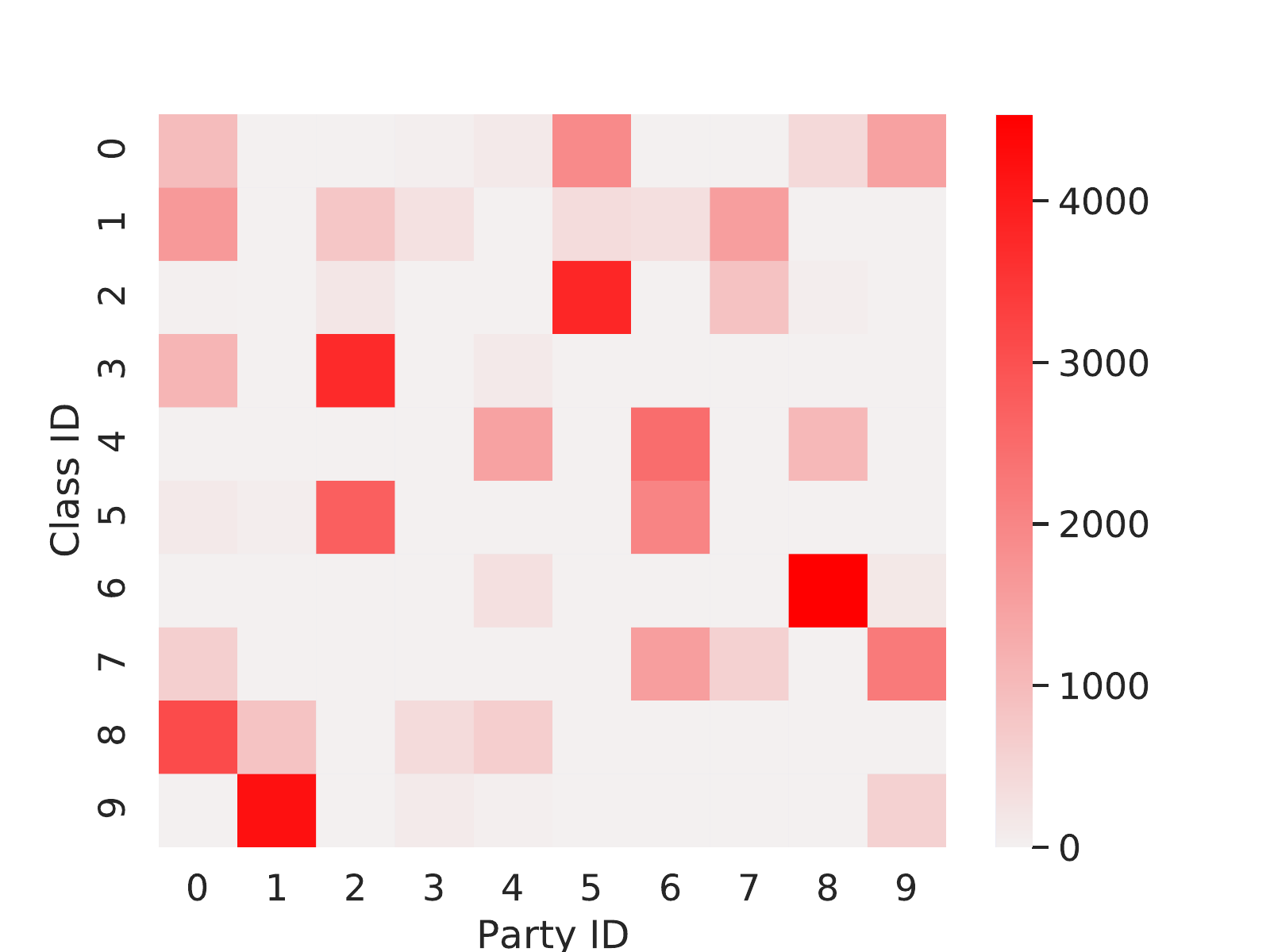}%
}
\hfill
% \hfil
\subfloat[CIFAR-100]{\includegraphics[width=0.32\textwidth]{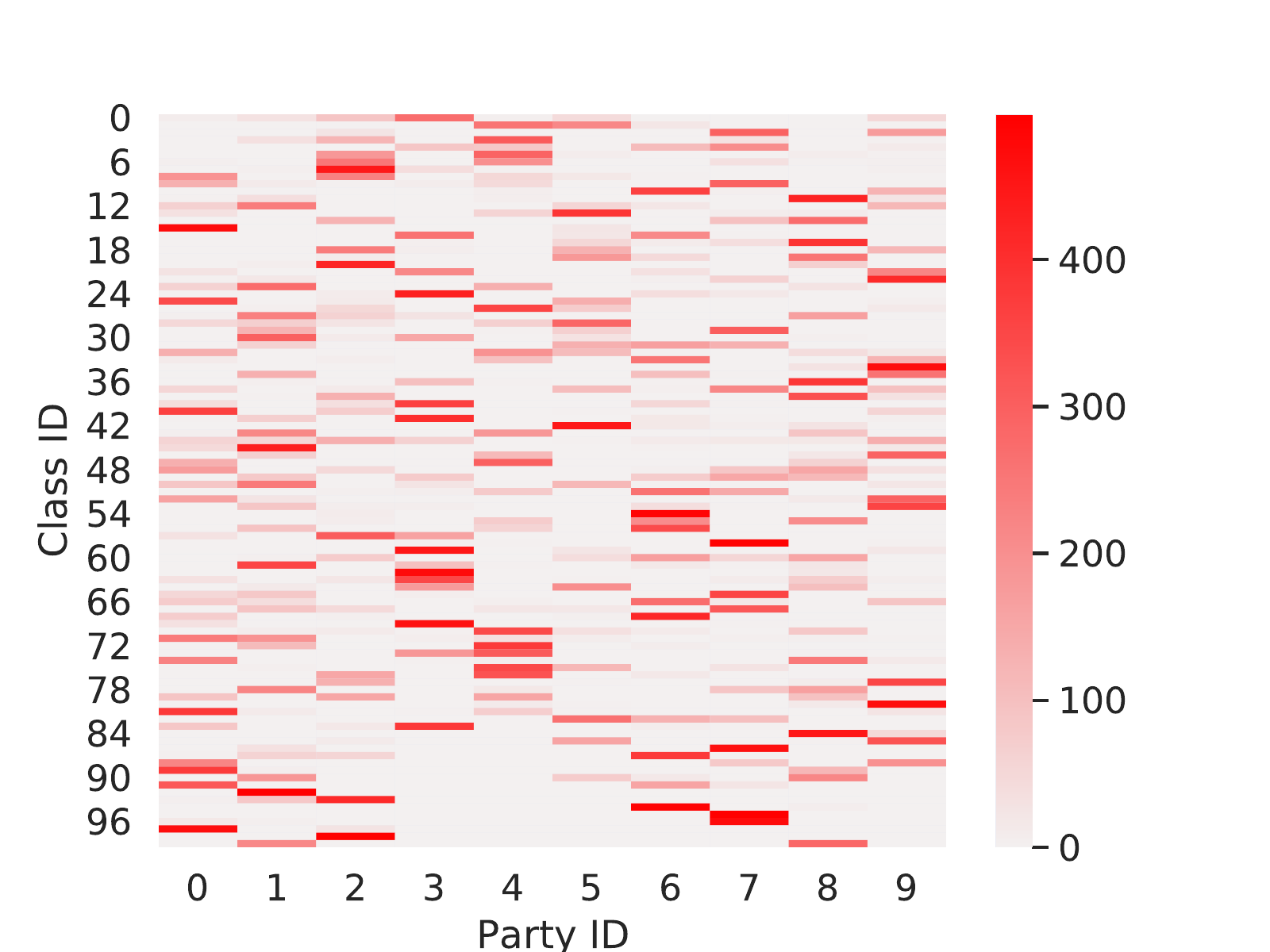}%
}
\hfill
\subfloat[Tiny-Imagenet]{\includegraphics[width=0.32\textwidth]{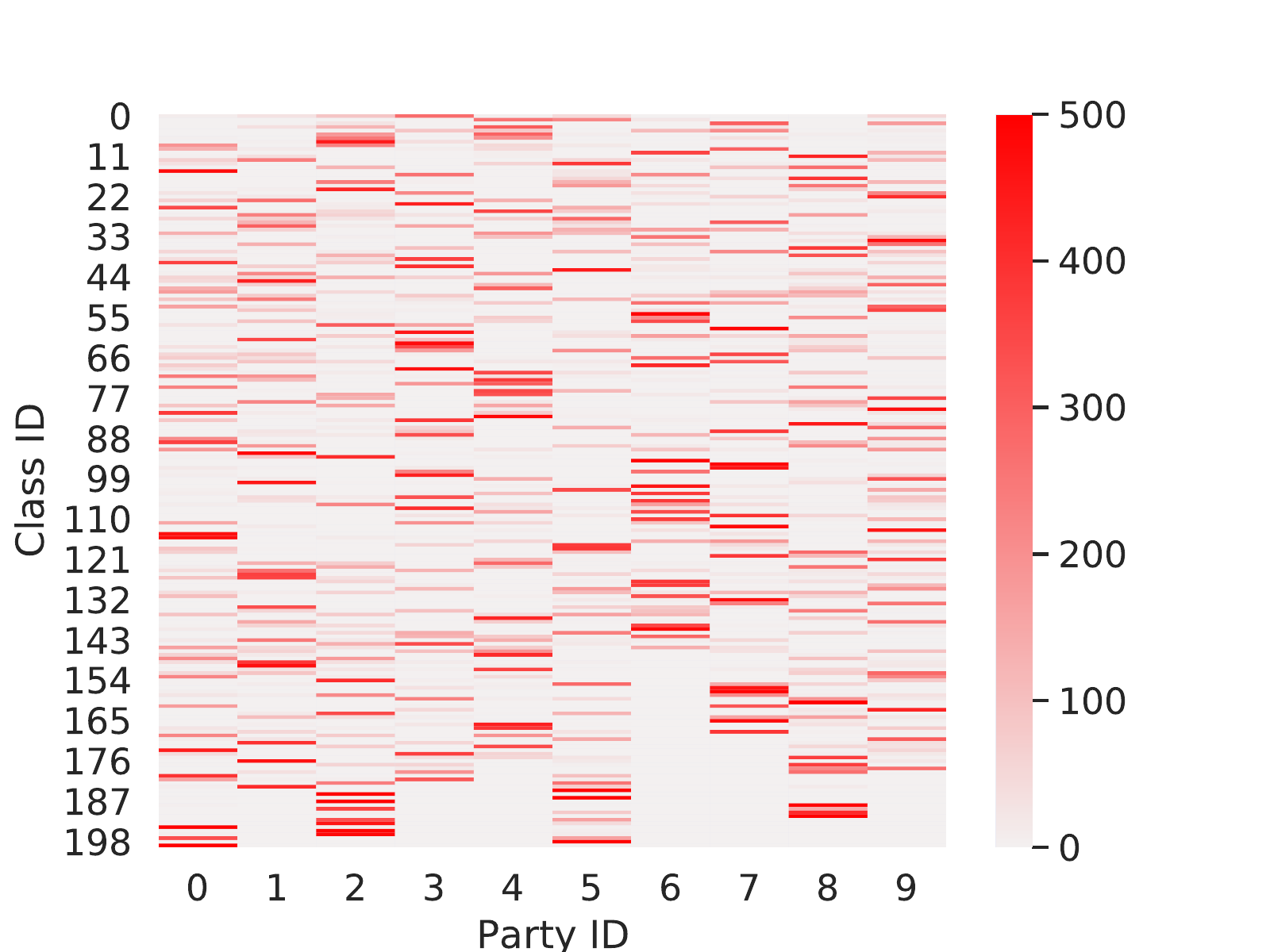}}
\caption{The data distribution of each party using non-IID data partition with $\beta=0.1$. }
\label{fig:datadis_beta01}
\vspace{-10pt}
\end{figure*}

\begin{figure*}
\centering
\subfloat[CIFAR-10]{\includegraphics[width=0.32\textwidth]{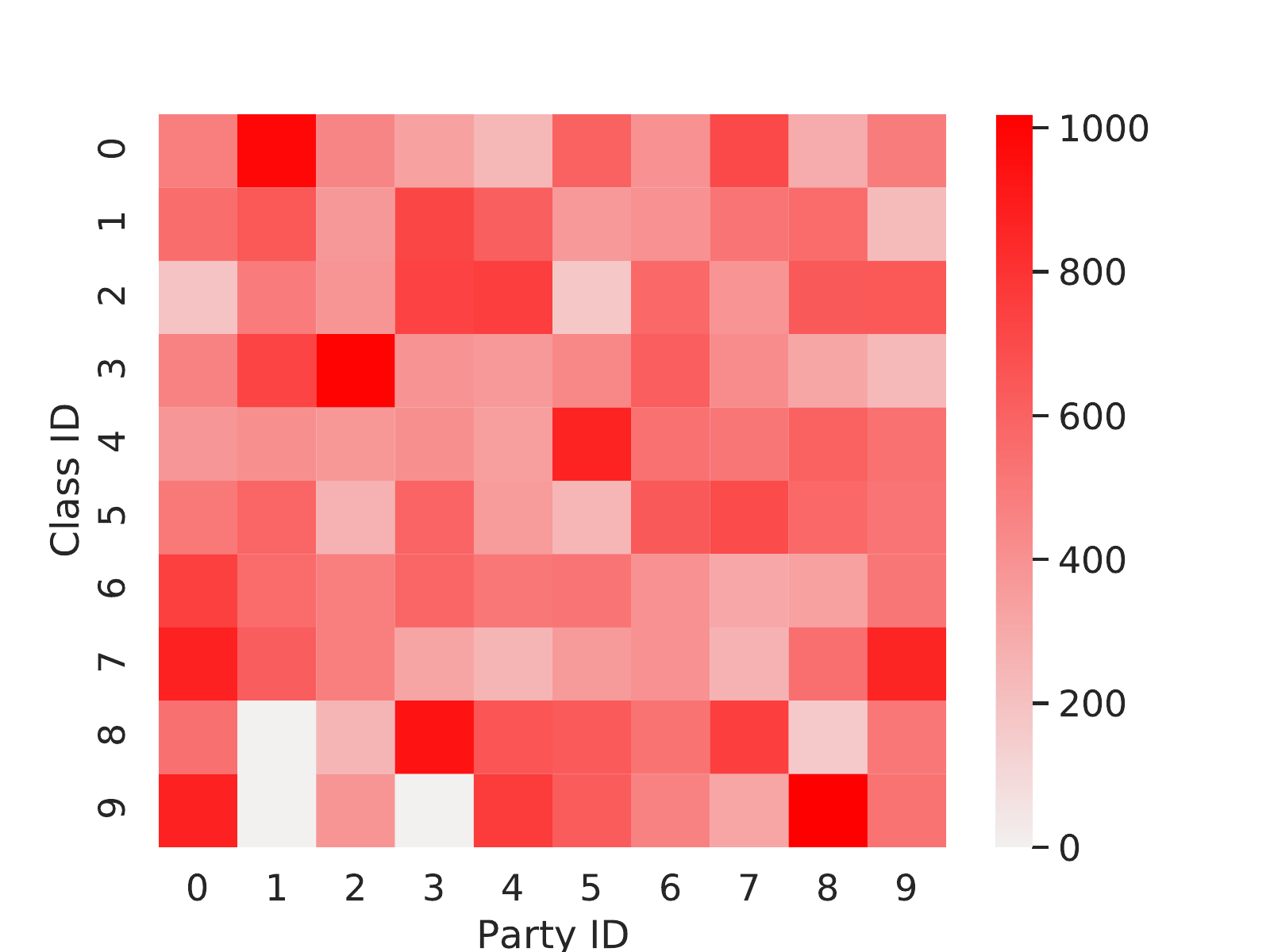}%
}
\hfill
% \hfil
\subfloat[CIFAR-100]{\includegraphics[width=0.32\textwidth]{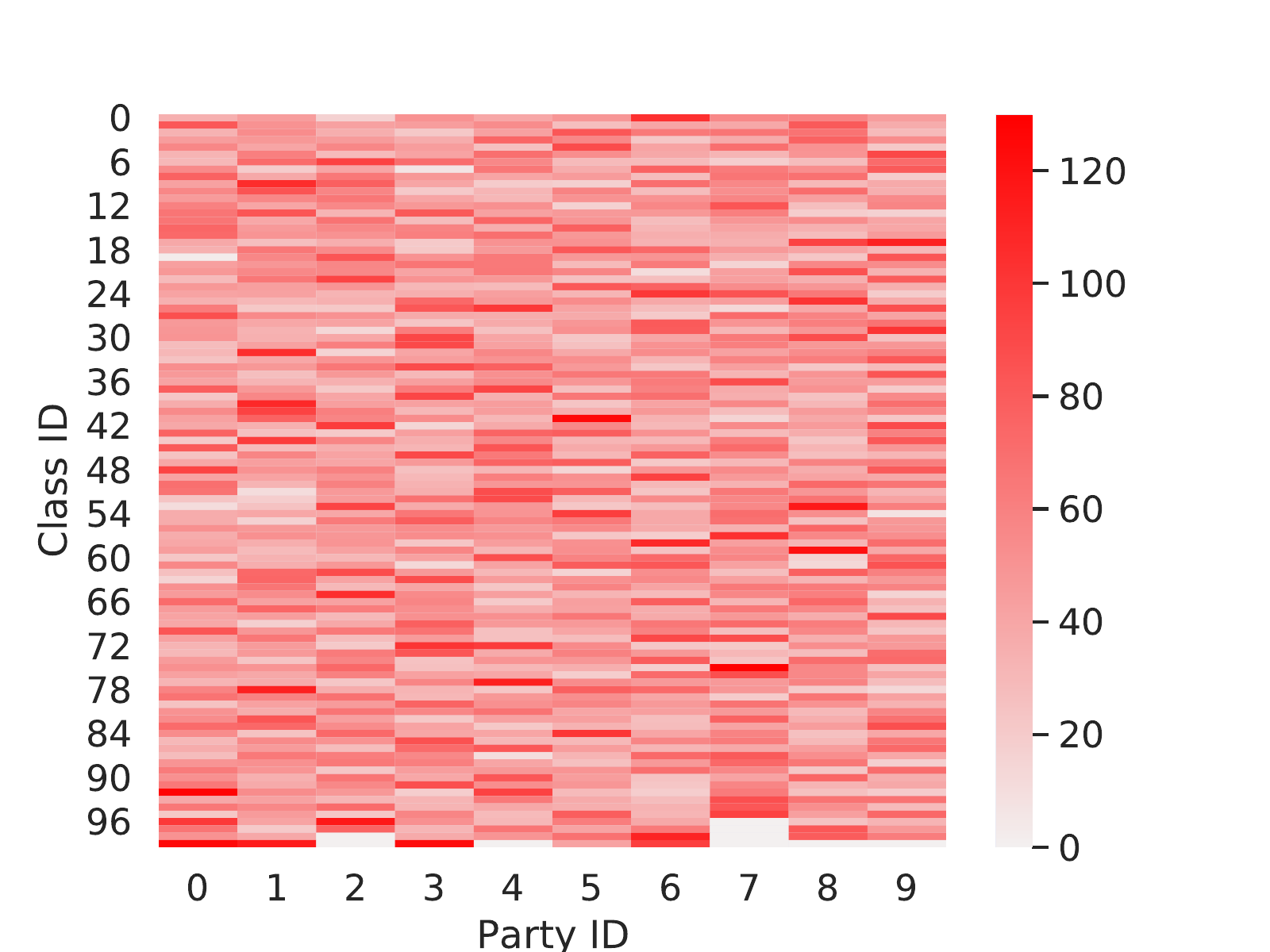}%
}
\hfill
\subfloat[Tiny-Imagenet]{\includegraphics[width=0.32\textwidth]{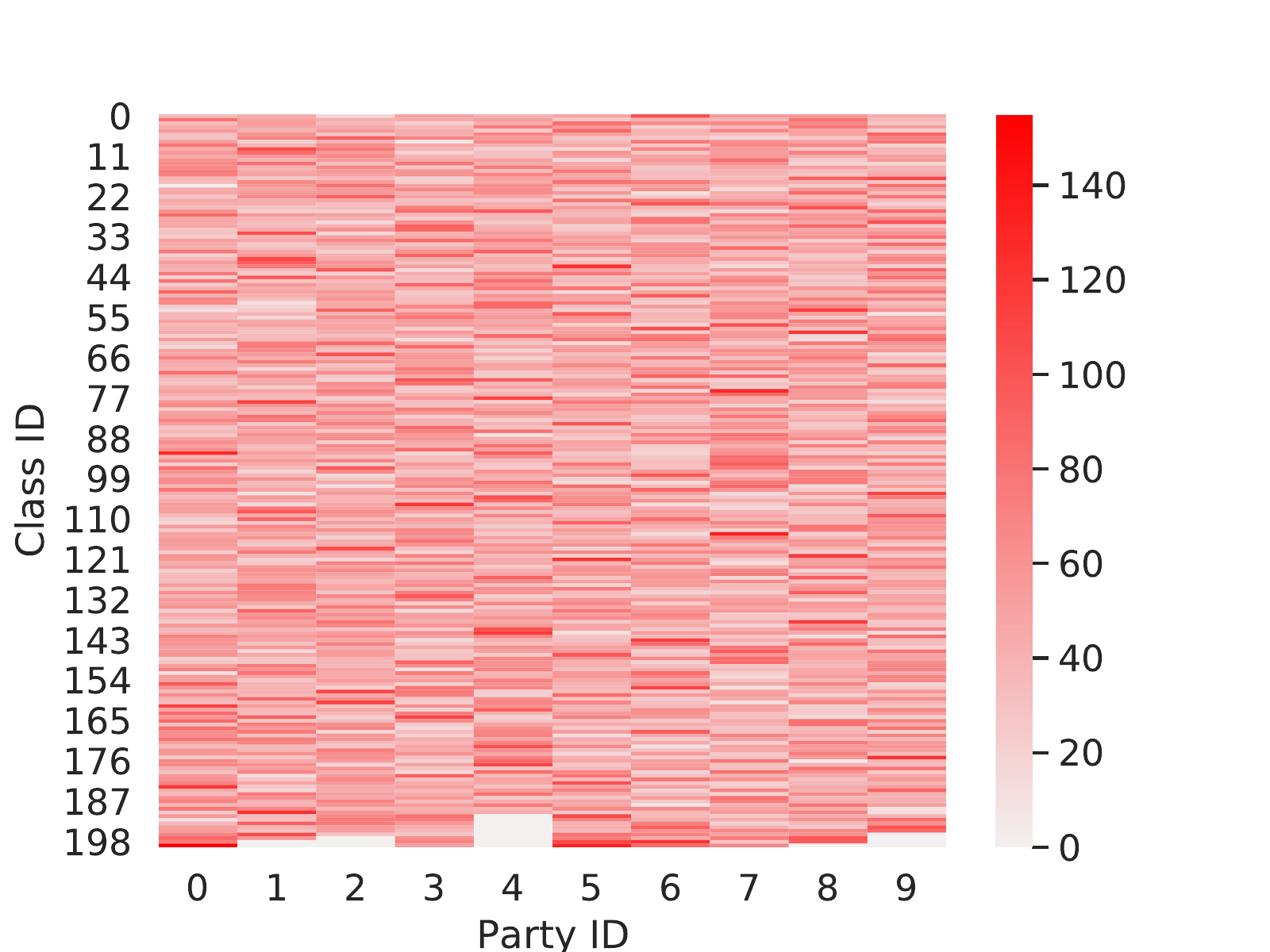}}
\caption{The data distribution of each party using non-IID data partition with $\beta=5$.}
\label{fig:datadis_beta5}
\vspace{-10pt}
\end{figure*}

Figure \ref{fig:datadis_beta01} and Figure \ref{fig:datadis_beta5} show the data distribution of $\beta=0.1$ and $\beta=5$ (used in Section 4.6 of the main paper), respectively.

\section{Projection Head}
We use a projection head to map the representation like \cite{simclr}. Here we study the effect of the projection head. We remove the projection head and conduct experiments on CIFAR-10 and CIFAR-100 (Note that the network architecture changes for all approaches). The results are shown in Table \ref{tbl:projhead}. We can observe that MOON can benefit a lot from the projection head. The accuracy of MOON can be improved by about 2\% on average with a projection head.

\begin{table}[h]
\centering
\caption{The top1-accuracy with/without projection head.}
\label{tbl:projhead}
\resizebox{\linewidth}{!}{
\begin{tabular}{|c|c|c|c|}
\hline
\multicolumn{2}{|c|}{Method} & CIFAR-10 & CIFAR-100 \\ \hline\hline
\multirow{5}{*}{\begin{tabular}[c]{@{}c@{}}without\\ projection \\ head\end{tabular}} & MOON & 66.8\% & 66.1\% \\ \cline{2-4} 
 & FedAvg & 66.7\% & 65.0\% \\ \cline{2-4} 
 & FedProx & 67.5\% & 65.4\% \\ \cline{2-4} 
 & SCAFFOLD & 67.1\% & 49.5\% \\ \cline{2-4} 
 & SOLO & 39.8\%$\pm$3.9\% & 22.5\%$\pm$1.1\% \\ \hline
\multirow{5}{*}{\begin{tabular}[c]{@{}c@{}}with\\  projection \\ head\end{tabular}} & MOON & \tb{69.1\%} & \tb{67.5\%} \\ \cline{2-4} 
 & FedAvg & 66.3\% & 64.5\% \\ \cline{2-4} 
 & FedProx & 66.9\% & 64.6\% \\ \cline{2-4} 
 & SCAFFOLD & 66.6\% & 52.5\% \\ \cline{2-4} 
 & SOLO & 46.3\%$\pm$5.1\% & 22.3\%$\pm$1.0\% \\ \hline
\end{tabular}
}
\end{table}

\section{IID Partition}
To further show the effect of our model-contrastive loss, we compare MOON and FedAvg when there is no heterogeneity among local datasets. The dataset is randomly and equally partitioned into the parties. The results are shown in Table \ref{tbl:homo}. We can observe that the model-contrastive loss has little influence on the training when the local datasets are IID. The accuracy of MOON is still very close to FedAvg even though with a large $\mu$. MOON is still applicable when there is no heterogeneity issue in data distributions across parties.

\begin{table}[h]
\centering
\caption{The top-1 accuracy of MOON and FedAvg with IID data partition on CIFAR-10.}
\label{tbl:homo}
% \resizebox{\linewidth}{!}{
\begin{tabular}{|c|c|c|}
\hline
\multicolumn{2}{|c|}{Method} & Top-1 accuracy\\ \hline\hline
\multirow{4}{*}{MOON} & $\mu=0.1$ & 73.6\%   \\ \cline{2-3} 
 & $\mu=1$ & 73.6\%   \\ \cline{2-3} 
 & $\mu=5$ & 73.0\%   \\ \cline{2-3} 
 & $\mu=10$ & 72.8\%   \\ \hline
\multicolumn{2}{|c|}{FedAvg ($\mu=0$)} & 73.4\%  \\ \hline
\end{tabular}
% }
\end{table}

\section{Hyper-Parameters Study}

\subsection{Effect of $\mu$}
We show the accuracy of MOON with different $\mu$ in Table \ref{tbl:mu}. The best $\mu$ for CIFAR-10, CIFAR-100, and Tiny-Imagenet are 5, 1, and 1, respectively. When $\mu$ is set to a small value (i.e., $\mu=0.1$), the accuracy of MOON is very close to FedAvg (i.e., $\mu=0$) since the impact of model-contrastive loss is small. As long as we set $\mu \geq 1$, MOON can benefit a lot from the model-contrastive loss. Overall, we find that $\mu=1$ is a reasonable good choice if they do not want to tune the parameter, where MOON achieves at least 2\% higher accuracy than FedAvg.

\begin{table}[!]
\centering
\caption{The test accuracy of MOON with $\mu$ from \{0, 0.1, 1, 5, 10\}. Note that MOON is actually FedAvg when $\mu=0$.}
\label{tbl:mu}
\begin{tabular}{|c|c|c|c|}
\hline
$\mu$ & CIFAR-10 & CIFAR-100 & Tiny-Imagenet \\ \hline\hline
0 & 66.3\% & 64.5\% & 23.0\% \\ \hline
0.1 & 66.5\% & 65.1\% & 23.4\% \\ \hline
1 & 68.4\% & \tb{67.5\%} & \tb{25.1\%} \\ \hline
5 & \tb{69.1\%} & 67.1\% & 24.4\% \\ \hline
10 & 68.3\% & 67.3\% & 25.0\% \\ \hline
\end{tabular}
\end{table}

\subsection{Effect of temperature and output dimension} 
\begin{figure}[h]
\centering
\subfloat[The effect of $\tau$\label{fig:tau}]{\includegraphics[width=0.48\linewidth]{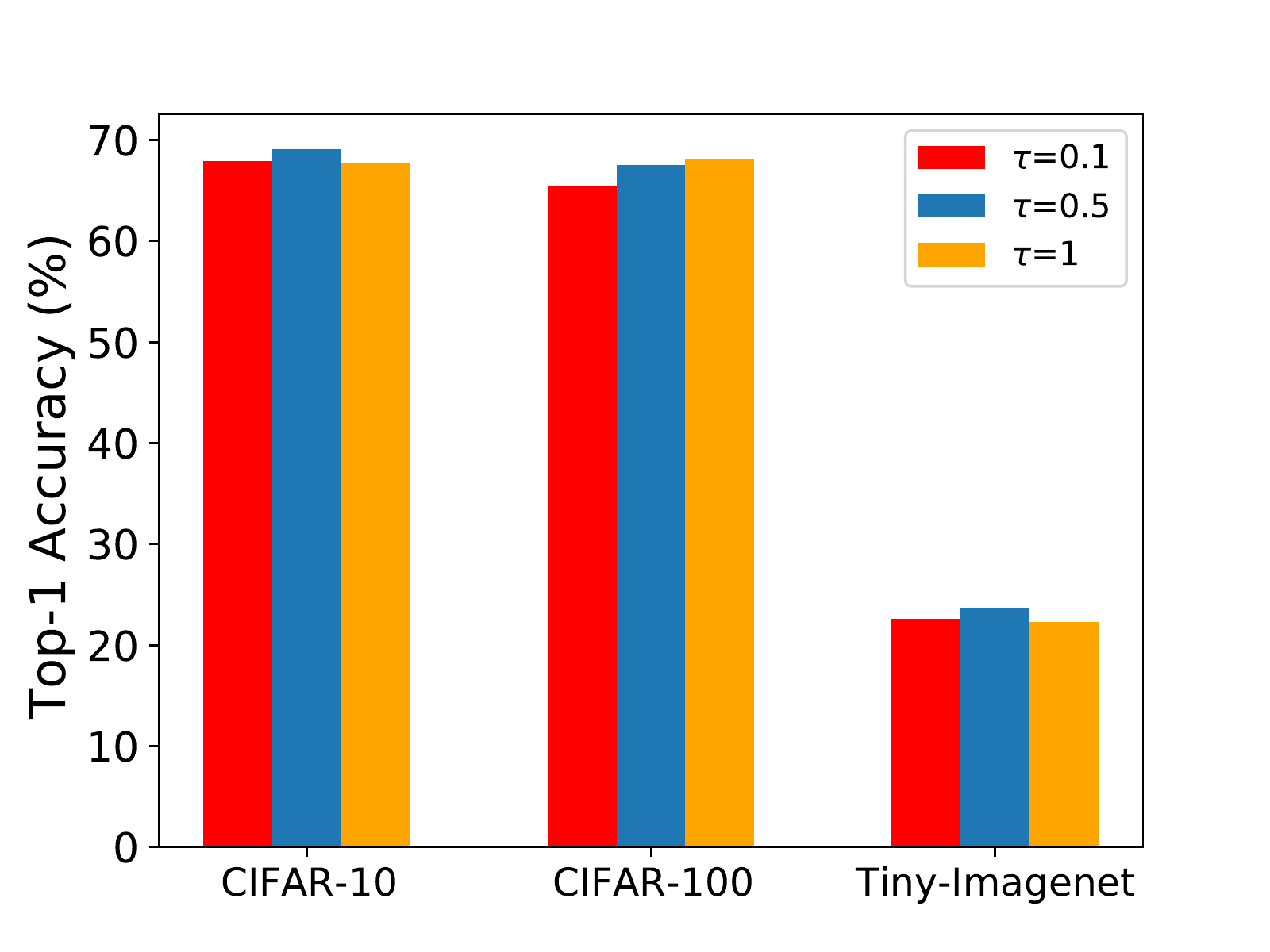}
}
\subfloat[The effect of output dimension\label{fig:outdim}]{\includegraphics[width=0.48\linewidth]{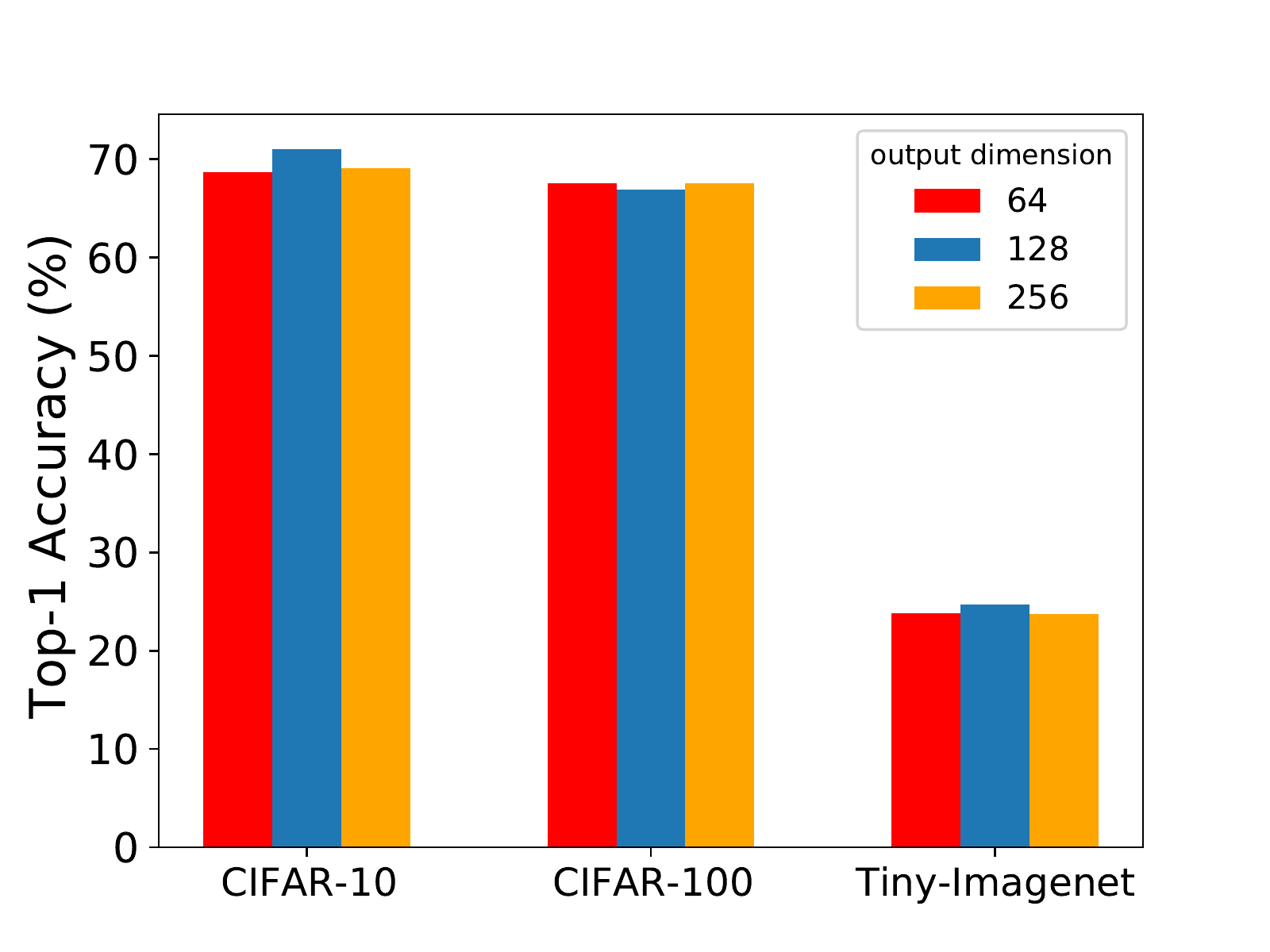}}
\caption{The top-1 accuracy of MOON trained with different temperatures and output dimensions.}
\label{fig:tau_outdim}
\vspace{-10pt}
\end{figure}

We tune $\tau$ from \{0.1, 0.5, 1.0\} and tune the output dimension of projection head from \{64, 128, 256\}. The results are shown in Figure \ref{fig:tau_outdim}. The best $\tau$ for CIFAR-10, CIFAR-100, and Tiny-Imagenet are 0.5, 1.0, and 0.5, respectively. The best output dimension for CIFAR-10, CIFAR-100, and Tiny-Imagenet are 128, 256, and 128, respectively. Generally, MOON is stable regarding the change of temperature and output dimension. As we have shown in the main paper, MOON already improves FedAvg a lot with a default setting of temperature (i.e., 0.5) and output dimension (i.e., 256). Users may tune these two hyper-parameters to achieve even better accuracy.

% \section{Real Federated Datasets}

\section{Combining with FedAvgM}
As we have mentioned in the fourth paragraph of Section 2.1, MOON can be combined with the approaches working on improving the aggregation phase. Here we combine MOON and FedAvgM \cite{hsu2019measuring}. We tune the server momentum $\beta \in \{0.1, 0.7, 0.9\}$. With the default experimental setting in Section 4.1, the results are shown in Table \ref{tbl:fedavgm}. While FedAvgM is better than FedAvg, MOON can further improve FedAvgM by 2-3\%.

\begin{table}[!]
\centering
\caption{The combining of MOON and FedAvgM.}
\label{tbl:fedavgm}
\resizebox{\columnwidth}{!}{
\begin{tabular}{|c|c|c|c|c|}
\hline
Datasets & FedAvg & MOON & FedAvgM & MOON+FedAvgM \\ \hline\hline
CIFAR-10 & 66.3\% & 69.1\% & 67.1\% & 69.6\% \\ \hline
CIFAR-100 & 64.5\% & 67.5\% &65.1\%  &67.8\%  \\ \hline
Tiny-Imagenet & 23.0\% & 25.1\% &23.4\%  &25.5\%  \\ \hline
\end{tabular}
}
\end{table}

\section{Computation Cost}
Since MOON introduces an additional loss term in the local training phase, the training of MOON will be slower than FedAvg. For the experiments in Table 1, the average training time per round with a NVIDIA Tesla V100 GPU and four Intel Xeon E5-2684 20-core CPUs are shown in Table \ref{tbl:local_time}. Compared with FedAvg, the computation overhead of MOON is acceptable especially on CIFAR-10 and CIFAR-100.

\begin{table}[!]
\centering
\caption{The average training time per round.}
\label{tbl:local_time}
\resizebox{\columnwidth}{!}{
\begin{tabular}{|c|c|c|c|}
\hline
Method & CIFAR-10 & CIFAR-100 & Tiny-Imagenet \\ \hline\hline
FedAvg & 330s & 20min & 103min \\ \hline
FedProx & 340s & 24min & 135min \\ \hline
SCAFFOLD & 332s & 20min & 112min \\ \hline
MOON & 337s & 31min & 197min \\ \hline
\end{tabular}
}
\vspace{-10pt}
\end{table}

\section{Number of Negative Pairs}
In typical contrastive learning, the performance usually can be improved by increasing the number of negative pairs (i.e., views of different images). In MOON, the negative pair is the local model being updated and the local model from the previous round. We consider using a single negative pair during training in the main paper. It is possible to consider multiple negative pairs if we include multiple local models from the previous rounds. Suppose the current round is $t$. Let $k$ denotes the maximum number of negative pairs. Let $z_{prev}^i = R_{w_i^{t-i}}(x)$ (i.e., $z_{prev}^i$ is the representation learned by the local model from $(t-i)$ round). Then, our local objective is

{\small
\begin{equation}
% \begin{aligned}
    \ell_{con}  = -\log{\frac{\exp(\textrm{sim}(z, z_{glob})/\tau)}{\exp(\textrm{sim}(z, z_{glob})/\tau) + \sum_{i=1}^k\exp(\textrm{sim}(z, z_{prev}^i)/\tau)}}
% \end{aligned}
\end{equation}
}

If $k=1$, then the objective is the same as MOON presented in the main paper. If $k>t$, since there are at most $t$ local models from previous rounds, we only consider the previous $t$ local models (i.e., only $t$ negative pairs). There is no model-contrastive loss if $t=0$ (i.e., the first round). Here we study the effect of the maximum number of negative pairs on CIFAR-10. The results are shown in Table \ref{tbl:n_pair}. Unlike typical contrastive learning, the accuracy of MOON cannot be increased by increasing the number of negative pairs. MOON can achieve the best accuracy when $k=1$, which is presented in our main paper.

\begin{table}[h]
\centering
\caption{The effect of maximum number of negative pairs. We tune $\mu$ from \{0.1, 1, 5, 10\} for all approaches and report the best accuracy.}
\label{tbl:n_pair}
\begin{tabular}{|c|c|}
\hline
maximum number of negative pairs & top-1 accuracy \\ \hline \hline
$k=1$ & \tb{69.1\%} \\ \hline
$k=2$ & 67.2\% \\ \hline
$k=5$ & 67.7\% \\ \hline
$k=100$ & 63.5\% \\ \hline
\end{tabular}
\end{table}

\end{document}